%% file: tell_draw_repeat.tex
\documentclass[10pt,twocolumn,letterpaper]{article}

\usepackage{iccv}
\usepackage{times}
\usepackage{epsfig}
\usepackage{graphicx}
\usepackage{amsmath}
\usepackage{amssymb}

\usepackage{glossaries}
\usepackage{tabu}
\usepackage{booktabs}
\usepackage{enumitem}
\usepackage{pifont}
\usepackage{authblk}
\usepackage{subcaption}
\usepackage[numbers]{natbib}

\usepackage[breaklinks=true,bookmarks=false]{hyperref}

\iccvfinalcopy

\def\iccvPaperID{6105}

\ificcvfinal\fi

\newacronym{cnn}{CNN}{Convolutional Neural Network}
\newacronym{mscoco}{MS COCO}{Microsoft Common Objects in Context}
\newacronym{gan}{GAN}{Generative Adversarial Network}
\newacronym{rnn}{RNN}{Recurrent Neural Network}
\newacronym{gru}{GRU}{Gated Recurrent Unit}
\newacronym{visdial}{VisDial}{Visual Dialog}
\newacronym{codraw}{CoDraw}{Collaborative Drawing}
\newacronym{clevr}{CLEVR}{Compositional Language and Elementary Visual Reasoning}
\newacronym{i-clevr}{i-CLEVR}{Iterative CLEVR}
\newacronym{geneva}{GeNeVA}{Generative Neural Visual Artist}
\newacronym{vqa}{VQA}{Visual Question Answering}
\newacronym{fid}{FID}{Fr\'echet Inception Distance}
\newacronym{is}{IS}{Inception Score}
\newacronym{ssm}{SSM}{Scene Similarity Metric}

\DeclareMathOperator\RelSim{rsim}
\newcommand\EGgen{\ensuremath{E_{G_{\text{gen}}}}}
\newcommand\EGgt{\ensuremath{E_{G_{\text{gt}}}}}
\newcommand{\cmark}{\ding{51}}
\newcommand{\xmark}{\ding{55}}

\begin{document}

\title{Tell, Draw, and Repeat: Generating and Modifying Images\\Based on Continual Linguistic Instruction}

\author{\vspace*{-5mm}
	Alaaeldin El-Nouby$^{1,4,}$\thanks{Work was performed during an internship with Microsoft Research.} \quad Shikhar Sharma$^{2}$ \quad Hannes Schulz$^{2}$ \quad Devon Hjelm$^{2,3,5}$ \\\vspace*{-3mm}
	Layla El~Asri$^{2}$ \quad Samira Ebrahimi~Kahou$^{2}$ \quad Yoshua Bengio$^{3,5,6}$ \quad Graham W.~Taylor$^{1,4,6}$ \\\vspace*{1mm}
	{\normalsize $^{1}$ University of Guelph \quad
	$^{2}$ Microsoft Research \quad
	$^{3}$ Montreal Institute for Learning Algorithms \\\vspace*{-1mm}
	$^{4}$ Vector Institute for Artificial Intelligence \quad
	$^{5}$ University of Montreal \quad
	$^{6}$ Canadian Institute for Advanced Research}}

\maketitle
\ificcvfinal\thispagestyle{empty}\fi

\vspace*{-10mm}
\begin{abstract}
   Conditional text-to-image generation is an active area of research, with many possible applications.
   Existing research has primarily focused on generating a single image from available conditioning information in one step.
   One practical extension beyond one-step generation is a system that generates an image iteratively, conditioned on ongoing linguistic input or feedback.
   This is significantly more challenging than one-step generation tasks, as such a system must understand the contents of its generated images with respect to the feedback history, the current feedback, as well as the interactions among concepts present in the feedback history.
   In this work, we present a recurrent image generation model which takes into account both the generated output up to the current step as well as all past instructions for generation.
   We show that our model is able to generate the background, add new objects, and apply simple transformations to existing objects. We believe our approach is an important step toward interactive generation.
   Code and data is available at: {\footnotesize \url{https://www.microsoft.com/en-us/research/project/generative-neural-visual-artist-geneva/}}.
\end{abstract}

\section{Introduction}
\begin{figure}[t!]
	\vspace*{-5mm}
	\centering
	\includegraphics[scale=0.42,trim={0 0 1.3cm 0},clip]{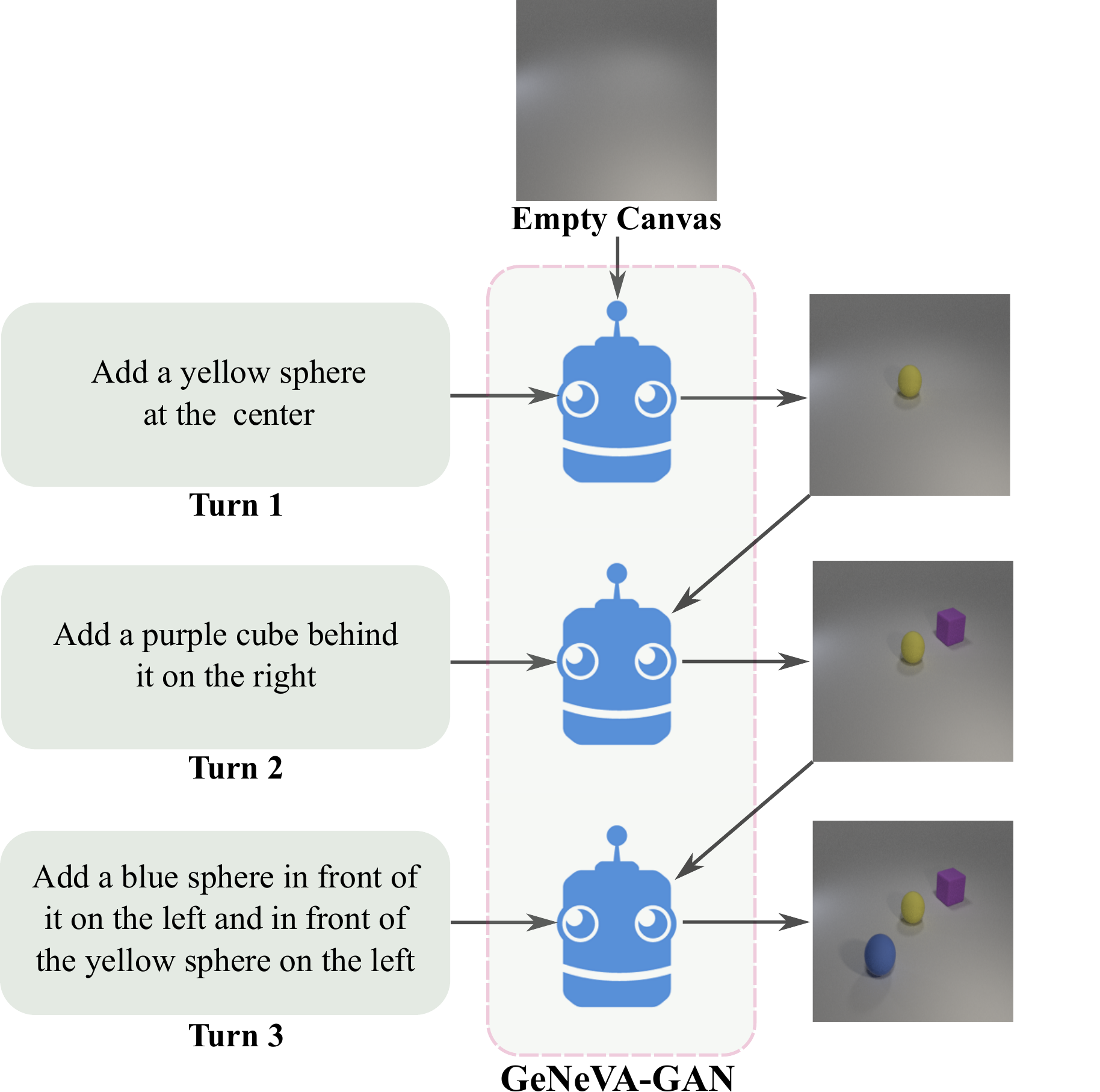}
	\caption{We present the Generative Neural Visual Artist (GeNeVA) task. Starting from an empty canvas, a \emph{Drawer} (GeNeVA-GAN) iteratively constructs a scene based on a series of instructions and feedback from a \emph{Teller}.}
   \label{fig:geneva}
   \vspace*{-4mm}
\end{figure}

\label{sec:introduction}
Vision is one of the most important ways in which humans experience, interact with, understand, and learn about the world around them.
Intelligent systems that can generate images and video for human users have a wide range of applications, from education and entertainment to the pursuit of creative arts.
Such systems also have the potential to serve as accessibility tools for the physically impaired; many modern and creative works are now generated or edited using digital graphic design tools, and the complexity of these tools can lead to inaccessibility issues, particularly with people with insufficient technical knowledge or resources.
A system that can follow speech- or text-based instructions and then perform a corresponding image editing task could improve accessibility substantially.
These benefits can easily extend to other domains of image generation such as gaming, animation, creating visual teaching material, etc.
In this paper, we take a step in this exciting research direction by introducing the \textit{neural visual artist} task.

Conditional generative models allow for generation of images from other input sources, such as labels~\cite{first_cgan} and dialogue~\cite{first_dialogue2image}. Image generation conditioned on natural language is a difficult yet attractive goal~\cite{first_caption2image,stackgan,honglak_layouts,attngan}.
Though these models are able to produce high quality images for simple datasets, such as birds, flowers, furniture, etc., good caption-conditioned generators of complex datasets, such as \gls{mscoco}~\cite{mscoco} are nonexistent.
This lack of good generators may be due to the limited information content of captions,
which are not rich enough to describe an entire image~\cite{first_dialogue2image}.
Combining object annotations with the intermediate steps of generating bounding boxes and object masks before generating the final images can improve results~\cite{honglak_layouts}.

Instead of constructing images given a caption, we focus on
learning to \emph{iteratively} generate images based on
continual linguistic input. We call this task the \gls{geneva}, inspired by
the process of gradually transforming a blank canvas to a scene.
Systems trained to perform this task should be
able to leverage advances in text-conditioned single image generation.

\subsection{GeNeVA Task and Datasets}
\label{geneva}

We present an example dialogue for the \gls{geneva} task in Figure~\ref{fig:geneva}, which involves a Teller giving a sequence of linguistic instructions to a Drawer for the ultimate goal of image generation.
The Teller is able to gauge progress through visual feedback of the generated image.
This is a challenging task because the Drawer needs to learn how to map complex linguistic instructions to realistic objects on a canvas, maintaining not only object properties but relationships between objects (e.g., relative location).
The Drawer also needs to modify the existing drawing in a manner consistent with previous images and instructions, so it needs to remember previous instructions.
All of these involve understanding a complex relationship between objects in the scene and how those relationships are expressed in the image in a way that is consistent with all instructions given.

For this task, we use the synthetic \gls{codraw} dataset~\cite{codraw}, which is composed of sequences of images along with associated dialogue of instructions and linguistic feedback (Figure~\ref{table:codraw_example}).
Also, we introduce the \gls{i-clevr} dataset (Figure~\ref{table:i-clevr_example}), a modified version of the \gls{clevr}~\cite{clevr} dataset, for incremental construction of \gls{clevr} scenes based on linguistic instructions.
Offloading the difficulty of generating natural images by using two well-studied synthetic datasets allowed us to better assess progress on the \gls{geneva} task and improve the iterative generation process. While photo-realistic images will undoubtedly be more challenging to work with, our models are by no means restricted to synthetic image generation. We expect that insights drawn from this setting will be crucial to success in the natural image setting.

\newcommand\dlgturn[2]{{\footnotesize {\textbf{#1:} #2}}}
\begin{figure*}[t]
   \small
   \begin{tabu} to \textwidth {@{}XXXX@{}}
   \centering \textbf{Turn 1} & \centering \textbf{Turn 2} & \centering
   \textbf{Turn 3} & \centering \textbf{Turn 4} \\
   \dlgturn{Teller}{top left corner big sun, orange part cut. right side far
   right medium apple tree. i see 4 apples} & \dlgturn{Teller}{left side girl big
   size, running, facing right. head above horizon.} & \dlgturn{Teller}{covering
   the tree, on the right side of the scene is a boy, kicking, facing left.
   head on green part. big size, black glasses. kicking ball.}&
   \dlgturn{Teller}{make tree a size bigger, move it up and left a bit. boys hand
   covers trunk.} \\
   \dlgturn{Drawer}{ok ready}& \dlgturn{Drawer}{ok}& \dlgturn{Drawer}{ok} &
   \dlgturn{Drawer}{ok}\\
   \centering \includegraphics[scale=0.15]{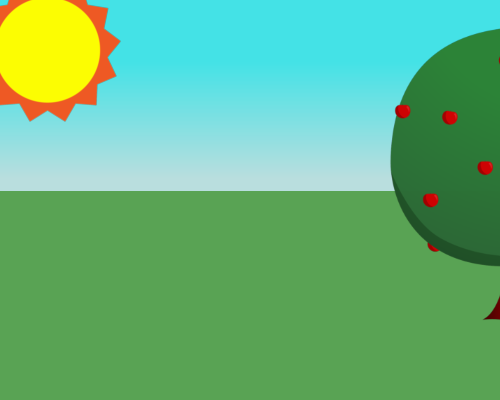} &
   \centering \includegraphics[scale=0.15]{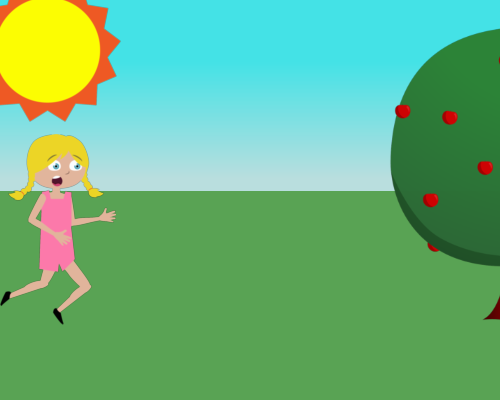} &
   \centering \includegraphics[scale=0.15]{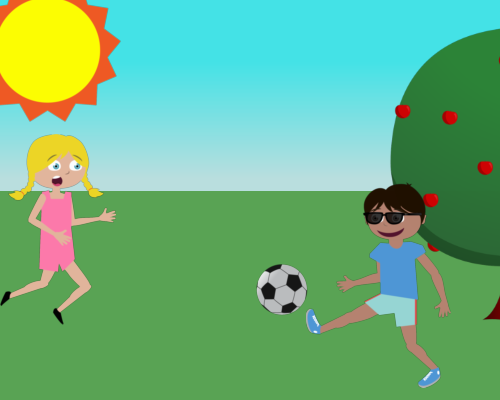} &
   \centering \includegraphics[scale=0.15]{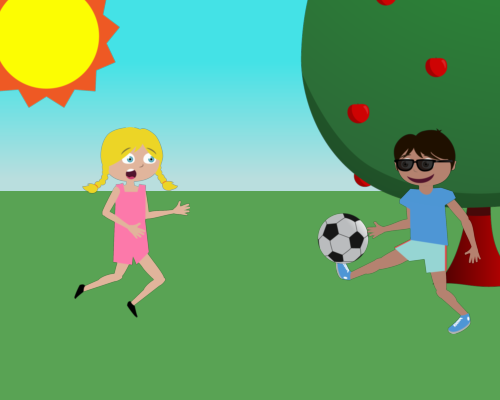} \\
   \end{tabu}
   \caption{An example from the \gls{codraw}~\cite{codraw} dataset. Each example from
      the dataset involves a conversation between a Teller and a Drawer. The Teller
      has access to a final image and has to iteratively provide text instructions
      and feedback to the Drawer to guide them to draw the same image. The Drawer
      updates the image on receiving instructions or feedback. In the original CoDraw
     setup, the Drawer predicted the position and attributes of objects to compose
     a scene. In GeNeVA, we task systems with generating the images directly in pixel space.}
   \label{table:codraw_example}
   \vspace*{-3mm}
\end{figure*}

The most similar task to \gls{geneva} is the task proposed
by the CoDraw \cite{codraw} authors. They require a model to build a scene
by placing the clip art images of the individual objects in
their correct positions. In other words, the model predictions
will be in coordinate space for their task, while for a
model for the \gls{geneva} task they will be in pixel space.
Natural images are in scope for the \gls{geneva} task, where \glspl{gan} are currently state-of-the-art.
Non-pixel-based approaches will be limited to placing pre-segmented specific poses of objects.
For such approaches, it will be extremely difficult to obtain
a pre-segmented set of all possible poses of all objects e.g., under
different lighting conditions. Additionally,
a pixel-based model does not necessarily require object-labels so
it can easily scale without such annotation.

\subsection{Contributions}
\label{contributions}

\noindent Our primary contributions are summarized as follows:
\begin{itemize}[noitemsep,nolistsep]
    \item We introduce the \gls{geneva} task and propose a novel recurrent
    \gls{gan} architecture that specializes in plausible modification of images in
    the context of an instructional history.
    \item We introduce the \gls{i-clevr} dataset, a sequential version of
    \gls{clevr}~\cite{clevr} with associated linguistic descriptions for constructing each \gls{clevr} scene, and establish a baseline for it.
    \item We propose a relationship similarity metric that evaluates the model's ability to place objects in a plausible position compatible with the instructions.
    \item We demonstrate the importance of iterative generation for complex scenes by showing that our approach outperforms the non-iterative baseline.
\end{itemize}

Our experiments on the \gls{codraw} and \gls{i-clevr} datasets show that our model is capable of generating images that incrementally build upon the previously generated images
and follow the provided instructions.
The model is able to learn complex behaviors such as drawing new objects, moving objects around in the image, and re-sizing these objects.
In addition to reporting qualitative results, we train an object localizer and measure precision, recall, F1 score, and our proposed relational similarity metric by comparing detections on ground-truth vs.~generated images.

\section{Related Work}
\label{sec:related_work}

\glspl{gan}~\cite{first_gan} represent a powerful family of generative models whose benefits and strengths extend to conditional image generation.
Several approaches for conditioning exist, such as conditioning both the generator and discriminator on
labels~\cite{first_cgan}, as well as training an auxiliary classifier as part of the discriminator~\cite{acgan}.
Closer to \gls{geneva} text-based conditioning,
\citet{first_caption2image} generate images conditioned on the provided captions.
\citet{stackgan} proposed a two-stage model called StackGAN, where the first stage generated low resolution images conditioned on the caption, and the second stage generated a higher resolution image conditioned on the previous image and the caption.
\citet{honglak_layouts} proposed a three-step generation process where they use external segmentation and bounding box annotation for \gls{mscoco} to first generate bounding boxes, then a mask for the object, and then the final image.
Building upon StackGAN, AttnGAN~\cite{attngan} introduced an attentional generator network that enabled the generator to synthesize different spatial locations in the image, conditioned on an attention mechanism over words in the caption.
It also introduced an image-text similarity module which encouraged generating images more relevant to the provided caption.

Departing from purely caption data, \citet{first_dialogue2image} proposed a non-iterative model called ChatPainter that generates images using dialogue data.
ChatPainter conditions on captions from \gls{mscoco} and a \gls{rnn}-encoded dialogue relevant to the caption (obtained from the \gls{visdial}~\cite{visdial} dataset) to generate images.
The authors showed that the question answering-based dialogue captured richer information about the image than just the caption, which enabled ChatPainter to generate superior images compared to using captions alone.
Since the \gls{visdial} dialogues were collected separately from the \gls{mscoco} dataset, there are no intermediate incremental images for each turn of the
dialogue.
The model, thus, only reads the entire dialogue and generates a single final image, so this setup diverges from a real-life sketch artist scenario where the artist has to keep making changes to the current sketch based on feedback.

There has also been recent work in performing recurrent image generation outside of text-to-image generation tasks.
\citet{lrgan} perform unsupervised image generation in recursive steps, first generating a background, subsequently conditioning on it to generate the foreground and the mask, and finally using an affine transformation to combine the foreground and background.
\citet{stgan} tackle the image compositing task of placing a foreground object on a background image in a natural location.
However, this approach is limited to fixed object templates, and instead of generating images directly, the model recursively generates parameters of transformations to continue applying to an object template until the image is close enough to natural image manifold.
Their approach also does not modify existing objects in the image.
Both of these approaches aim to generate a single final image without incorporating any external feedback.
To the best of our knowledge, the proposed model is the first of its kind that can recursively generate and modify intermediate images based on continual text instructions such that every generated image is consistent with past instructions.

\section{Methods}
\label{sec:methods}

\begin{figure*}
    \centering
    \includegraphics[scale=0.76]{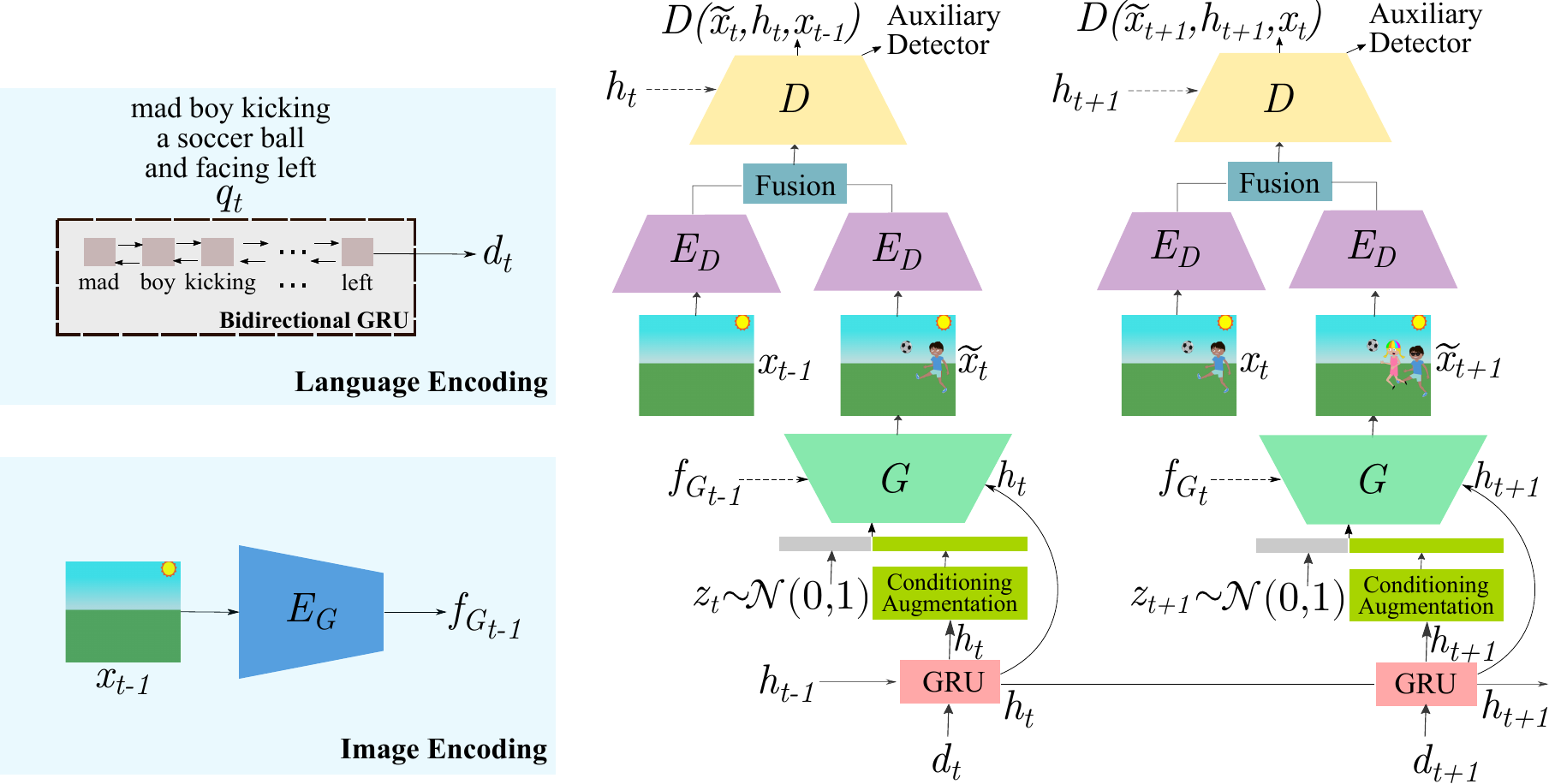}
    \caption{Overview of the \gls{geneva}-\gls{gan} architecture. For each time-step $t$, instruction $q_{t}$ is encoded into $d_{t}$ using a bi-directional GRU. The previous time-step generated image $\tilde{x}_{t-1}$ (teacher-forcing at training time with ground truth $x_{t-1}$) is encoded into $f_{G_{t-1}}$ using $E_{G}$. A GRU outputs a context-aware condition $h_{t}$ as a function of $d_{t}$ and the previous condition $h_{t-1}$. The generator $G$ generates an image $\tilde{x}_{t}$ conditioned on $h_{t}$ and $f_{G_{t-1}}$. $f_{G_{t-1}}$ is concatenated to feature maps from $G$ with the same spatial dimensions while $h_{t}$ is used as the input for conditional batch normalization. The image from the current time-step (ground truth $x_{t}$ or generated $\tilde{x}_{t}$) and the previous time-step ground-truth image are encoded using $E_D$. The features from both images are fused and then passed as input to a discriminator $D$. Finally, $D$ is conditioned using the context-aware condition $h_{t}$. An auxiliary objective of detecting all the objects in the scene is also added to $D$.}
	\label{fig:model}
	\vspace*{-3mm}
\end{figure*}

In this section, we describe a conditional recurrent \gls{gan} model for the
\gls{geneva} task.
An overview of the model architecture is shown in Figure~\ref{fig:model}.

\subsection{Model}
\vspace*{-1mm}
During an $n$-step interaction between a Teller and a Drawer, the Teller provides a drawing
canvas $x_{0}$ and a sequence of instructions $Q=(q_{1},\dotsc, q_{n})$. For
every turn in the conversation, a conditioned generator $G$ outputs a new
image
\vspace*{-5mm}
\begin{align}
	\widetilde{x}_{t} = G(z_{t}, h_{t}, f_{G_{t-1}}) ,\
	\label{eq:gen}
\end{align}
\vspace*{-6mm}\\
where $z_{t}$ is a noise vector sampled from a normal distribution
$\mathcal{N}(0, 1)$ of dimension $N_{z}$. $G$ is conditioned on two variables,
$h_{t}$ and $f_{G_{t-1}}$, where $h_{t}$ is a context-aware condition and
$f_{G_{t-1}}$ is context-free.

The context-free condition $f_{G_{t-1}} = E_{G}(\widetilde{x}_{t-1})$ is an encoding of the previously
generated image $\widetilde{x}_{t-1}$ using an encoder $E_{G}$, which is a
shallow \gls{cnn}. Assuming square inputs, the encoder produces low resolution
feature maps of dimensions ($K_{g} {\times} K_{g} {\times} N_{g}$).

The context-aware condition $h_t$ needs to have access to the conversation
history such that it can learn a better encoding of the instruction in the
context of the conversation history up to time $t-1$.

Each instruction $q_{t}$ is encoded using a bi-directional \gls{gru}
on top of GloVe word embeddings \cite{glove}. This instruction encoding
is denoted by $d_{t}$.

We formulate $h_{t}$ as a recursive function $R$, which takes
the instruction encoding $d_{t}$ as well as
the previous condition $h_{t-1}$ as inputs. We implement $R$ with a second
\gls{gru}, which yields $h_{t}$ with dimension $N_{c}$:
\vspace*{-2mm}
\begin{align}
	h_{t} = R(d_{t}, h_{t-1}).\
	\label{eq:R}
\end{align}
\vspace*{-10mm}\\

The context-free condition $f_{G_{t-1}}$ represents the prior given to the
model by the most recently generated image (i.e.~a representation of the
current canvas). On the other hand, the context-aware
condition $h_t$ represents the modifications the Teller is
describing in the new image. In our model, the context-aware condition is concatenated with
the noise vector $z_{t}$ after applying conditioning augmentation~\cite{stackgan}, as
shown in Figure~\ref{fig:model}. Similar to \citet{cgan_w_proj}, it
is also used in applying conditional batch normalization to all of the generator's
convolutional layers. The context-free condition $f_{G_{t-1}}$ is concatenated with the feature maps from the generator's intermediate layer $L_{f_{G}}$ which has the same spatial dimensions as $f_{G_{t-1}}$.

Since we are modeling iterative modifications of images, having a discriminator
$D$ that only distinguishes between real and generated images at each step will not
be sufficient. The discriminator should also identify cases where the image is
modified incorrectly with respect to the instruction or not modified at all.
To enforce this, we introduce three modifications
to the discriminator. First, an image encoder $E_{D}$ is used to encode the current time step image (real or generated) \emph{and the previous time-step ground-truth image} as shown in Figure~\ref{fig:model}.  The output feature maps of dimensions ($K_{d} {\times}
K_{d}{\times} N_{d}$) are passed through a fusion layer. We experiment with element-wise subtraction and concatenation of feature maps as different options for fusion. The fused features are passed through a discriminator $D$. Passing a fused representation of both the current and the previous images to the discriminator encourages it to focus on the quality of the modifications, not only the overall image quality. This provides a better training signal for the generator.
 Additionally, the context-aware condition $h_{t-1}$ is used as a condition for $D$ through projection similar to \cite{cgan_w_proj}.

 Second, for the discriminator loss, in addition to labelling real images as
 positive examples and generated images as negative examples, we add a term for
 the combination of [real image, wrong instruction], similar to
 \citet{first_caption2image}. Finally, we add an auxiliary objective \cite{acgan} of detecting all objects in the scene at the current time step.

 The generator and discriminator are trained alternately to minimize the
 adversarial hinge loss \cite{geo_gan, sn_gan, sagan}. The discriminator minimizes
 \vspace*{-2mm}
 \begin{align}
 L_{D} = L_{D_{\text{real}}} + \frac{1}{2} (L_{D_{\text{fake}}} + L_{D_{\text{wrong}}} )+ \beta L_{\text{aux}} ,\
 \label{eq:D}
 \end{align}
 \vspace*{-7mm}\\
 where
 \vspace*{-1mm}
 \begin{align*}
 L_{D_{\text{real}}} &= - \mathbb{E}_{(x_{t}, c_{t})\sim p_{\text{data}(0:T)}} [\min(0, -1 + D(x_{t}, c_{t}))] \\
 L_{D_{\text{wrong}}} &= - \mathbb{E}_{(x_{t}, \hat{c_{t}})\sim p_{\text{data}(0:T)}}  [\min(0, -1 - D(x_{t}, \hat{c_{t}}))]\\
 L_{D_{\text{fake}}} &= - \mathbb{E}_{z_{t}\sim \mathcal{N}, c_{t} \sim p_{\text{data}(0:T)}} [\min(0, -1 - D(G(z_{t}, \tilde{c}_{t}), c_{t}))] ,\
 \end{align*}
 \vspace*{-5mm}\\
with $\tilde{c}_{t}= \{h_{t}, f_{G_{t-1}}\}$ and $c_{t}=\{h_{t}, x_{t-1}\}$. Finally, $\hat{c_{t}}$ is the same as $c_{t}$ but with a wrong instruction and $T$ is the length of the instruction sequence $Q$.

 The loss function for the auxiliary task is a binary cross entropy over all the $N$
 possible objects at that time step,
 \vspace*{-2mm}
 \begin{align*}
     L_{\text{aux}} = \sum_{i=0}^{N} - \left(y_{i} \log(p_{i}) + (1-y_{i}) \log(1-p_{i})\right) ,\
 \end{align*}
 \vspace*{-4mm}\\
 where $y_{i}$ is a binary label for each object indicating whether it is present in the scene at the current time step. Note that we do not index the loss with $t$
 to simplify notation.
 A linear layer of dimension $N$ is added to the last discriminator layer before applying projection conditioning with $h_{t}$. A sigmoid function is applied to each of the $N$ outputs yielding $p_{i}$,
 the model detection prediction for object $i$.

 The generator loss term is
 \vspace*{-1mm}
 \begin{align}
  \hspace*{-2mm}L_{G} =  - \mathbb{E}_{z_{t} \sim p_{z}, c_{t} \sim p_{data}(0:T)} D(G(z_{t}, \tilde{c}_{t}), c_{t}) + \beta L_{\text{aux}} \
  \label{eq:G}
 \end{align}
 \vspace*{-8mm}\\

 Additionally, to help with training stability, we apply zero-centered gradient
 penalty regularization to the discriminator's parameters $\Phi$ on the real data alone with weighting factor $\gamma$ as suggested by \citet{MeschederICML2018},
 \vspace*{-2mm}
 \begin{align}
    \text{GPReg}(\Phi)  = \frac{\gamma}{2} \mathbb{E}_{p_{D(x)}} [\|{\nabla D_{\Phi}(x)}\|^{2}].
 \end{align}

\subsection{Implementation Details}
The network architecture for the generator and discriminator follows the
ResBlocks architecture as used by \citet{cgan_w_proj}. Following SAGAN
\cite{sagan}, we add a self-attention layer to the intermediate layers with
spatial dimensions of $16 {\times} 16$ for the discriminator and the generator. We use spectral normalization \cite{sn_gan} for all layers in the
discriminator.

For the training dynamics, the generator and discriminator
parameters are updated every time step, while the parameters of $E_{G}$, $R$ and the text encoder are
updated every sequence. The text encoder and the network $R$ are
trained with respect to the discriminator objective only.

We add layer normalization \cite{layernorm} to the text encoding \gls{gru}, as well
as the the \gls{gru} implementing $R$. We add batch normalization \cite{batchnorm} to the
output of the image encoder $E_{G}$. We found that adding these normalization
methods was important for gradient flow to all modalities.

For training, we used teacher forcing by using the ground truth images $x_{t-1}$
instead of the generated image $\widetilde{x}_{t-1}$, but we use
$\widetilde{x}_{t-1}$ during test time. We use the Adam~\cite{adam} optimizer
for the GAN, with learning rates of 0.0004 for the discriminator and
the 0.0001 for the generator, trained with an equal number of updates. We use Adam as well for
the text encoder with learning rate of 0.003, and for the GRU with
learning rate of $3\cdot 10^{-4}$.

In our experiments the following hyper-parameters worked the best, $N_{z}=100$, $N_{c}=1024$,  $K_{g}=16$, $N_{g}=128$, $K_{d}=16$, $N_{d}=256$, $\gamma=10$, and $\beta=20$. More details are provided in the appendix.

\section{Datasets}
\label{sec:datasets}
For the \gls{geneva} task, we require a dataset that contains textual
instructions describing drawing actions, along with corresponding ground truth
images for each instruction. To the best of our knowledge, the only such dataset
publicly available is \gls{codraw}. Additionally, we
create a new dataset called \gls{i-clevr}, specifically designed for this task.

\subsection{CoDraw}
\gls{codraw} \cite{codraw} is a recently released clip art-like dataset. It
consists of scenes, which are sequences of images of children playing in a park. The children have
different poses and expressions and the scenes include other objects such as trees,
tables, and animals. There are 58 object types in total. Corresponding to every scene, there
is a conversation between a Teller and a Drawer (both Amazon Mechanical Turk workers) in
natural language. The Drawer updates the canvas based on the Teller's instructions.
The Drawer can ask questions as well for clarification. The dataset consists of
9,993 scenes of varying length. An example of such a scene is shown in
Figure~\ref{table:codraw_example}. The initial drawing canvas $x_0$ for
\gls{codraw} provided to the Drawer consists of the background having just the
sky and grass.

\vspace*{1mm}
\noindent \textbf{Pre-processing:} In some instances of the original dataset, the Drawer
waited for multiple Teller turns before modifying the image. In these cases, we
concatenate consecutive turns into a single turn until the Drawer modifies the
image. We also concatenate turns until a new object has been added or removed.
Thus every turn has an image in which the number of objects has changed since the last turn.

We treat the concatenated utterances of the Drawer and the Teller at
time step $t$ as the instruction, injecting a special delimiting token between the Teller and Drawer. The Teller and Drawer text contains several spelling
mistakes and we run the Bing Spell Check API\footnote{\scriptsize \url{https://azure.microsoft.com/en-us/services/cognitive-services/spell-check/}}
over the entire dataset to make corrections. For words that are not present in the
GloVe vocabulary, we use the ``unk'' word embedding from GloVe. We use the
same train-valid-test split proposed in the original \gls{codraw} dataset.

\subsection{\gls{i-clevr}}
\newcommand\iclvrex[1]{{\footnotesize #1}}
\begin{figure*}[t]
	\small
	\begin{tabu} to \textwidth {XXXXX}
		\centering \includegraphics[scale=0.23]{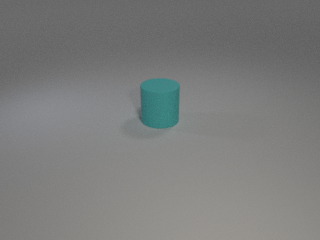} &
		\centering \includegraphics[scale=0.23]{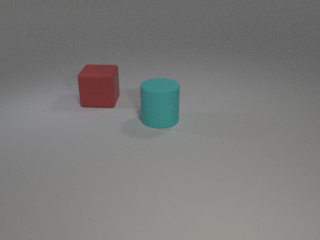} &
		\centering \includegraphics[scale=0.23]{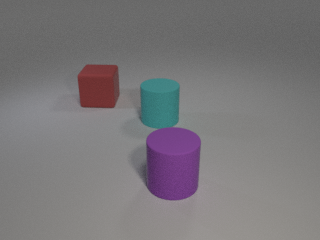} &
		\centering \includegraphics[scale=0.23]{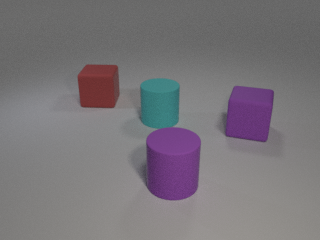} &
		\centering \includegraphics[scale=0.23]{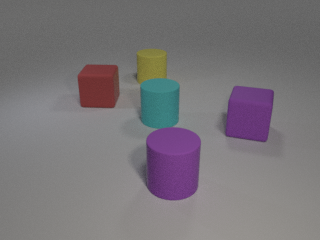}
		\\
		\iclvrex{Add a cyan cylinder at the center} &
		\iclvrex{Add a red cube behind it on the left} &
		\iclvrex{Add a purple cylinder in front of it on the right and in front of the cyan cylinder} &
		\iclvrex{Add a purple cube behind it on the right and in front of the red cube on the right} &
		\iclvrex{Add a yellow cylinder behind the purple cylinder on the left and behind the red cube on the right}
	\end{tabu}
	\vspace*{-1mm}
	\caption{Example sequence of image-instruction pairs from the \gls{i-clevr} dataset.}
	\label{table:i-clevr_example}
	\vspace*{-4mm}
\end{figure*}

\gls{clevr} \cite{clevr} is a programmatically generated dataset that is popular in the \gls{vqa} community. \gls{clevr} consists of images of
collections of objects with different shapes, colors, materials and sizes. Each
image is assigned complex questions about object counts, attributes or existence.
We build on top of the open-source generation
code\footnote{\scriptsize \url{https://github.com/facebookresearch/clevr-dataset-gen}} for
\gls{clevr} to create Iterative \gls{clevr} (\gls{i-clevr}). Each example in
the dataset consists of a sequence of 5 (image, instruction) pairs. Starting
from an empty canvas (background), each instruction describes an object to add
the canvas in terms of its shape and color. The instruction also describes where the
object should be placed relative to existing objects in the scene. To make the
task more complex and force the model to make use of context, we refer to the most
recently added object by ``it'' instead of stating its attributes. An example from
the \gls{i-clevr} dataset is presented in Figure~\ref{table:i-clevr_example}. The
initial drawing canvas $x_0$ for \gls{i-clevr} consists of the empty background. A
model is tasked with learning how to add the object with the correct attributes
in a plausible position, based on the textual instruction. More details about the dataset generation can be found in the appendix.

The i-CLEVR dataset consists of 10,000 sequences, totalling 50,000 images
and instructions. The training split contains 6,000 sequences, while the
validation and testing splits have 2,000 sequences each.
\begin{table*}[!t]
	\small
	\centering
	\begin{tabu}{@{}X[l,1.5,b]*3{X[c,1,b]}X[c,1.9,b]*3{X[c,1,b]}X[c,1.9,b]@{}}
		\toprule
		&  \multicolumn{4}{c}{CoDraw} &  \multicolumn{4}{c}{i-CLEVR} \\
		\cmidrule(r){2-5}
		\cmidrule(l){6-9}
		Model & Precision & Recall & F1-Score & $\RelSim(\EGgt,\EGgen)$ & Precision & Recall & F1-Score & $\RelSim(\EGgt,\EGgen)$\\
		\midrule
		Non-iterative & 50.60 & 43.42 & 44.96 & 22.33 & 25.49 & 20.95 & 22.63 & 11.52 \\ \midrule
		Baseline & 55.61 & 42.31 & 48.05 & 25.31 & 69.09 & 56.38 & 62.08 & 45.19 \\
		Mismatch & 62.47 & 48.95 & 54.89 & 32.74 & 71.15 & 60.57 & 65.44 & 50.21 \\
		$G$ prior  & 60.78 & 49.37 & 54.48 & 33.60 & 82.80 & 77.22 & 79.91 & 63.93 \\
		Aux &  54.78 & 51.51 & 53.10 & 33.83 & 83.63 & 75.63 & 79.43 & 55.36 \\
		$D$ Concat &  66.38 & 51.27 & 57.85 & 33.57 & 88.47 & 83.35 & 85.83 & 70.22 \\
		$D$ Subtract & \textbf{66.64} & \textbf{52.66} & \textbf{58.83} & \textbf{35.41} & \textbf{92.39} & \textbf{84.72} & \textbf{88.39} &\textbf{ 74.02 }\\  \bottomrule
   \end{tabu}
   \caption{Results of the GeNeVA-GAN ablation study on the CoDraw and i-CLEVR datasets.}
\label{table:ablation}
\vspace*{-5mm}
\end{table*}

\section{Experiments}
\label{sec:experiments}

In this section, we first define our evaluation metrics, and then describe the experiments
carried out on the \gls{codraw} and \gls{i-clevr} datasets.

\subsection{Evaluation Metrics}
Standard metrics used for evaluating GAN models such as the Inception Score or \gls{fid}
only capture how realistic the generations look relative to real images. They
cannot detect if the model is correctly modifying the images according to
the \gls{geneva} task instructions. A good evaluation metric for this task needs
to identify if all the objects that were described by the Teller are present in
the generated images. It should also check that the objects' positions and
relationships match the instructions. To capture all of these constraints, we
train an object localizer on the training dataset. For every example, we compare
the detections of this localizer on the real images and the generated ones. We
present the precision, recall, and F1-score for this object detection task. We
also construct a graph where the nodes are objects present in the images and
edges are positional relationships: left, right, behind, front. We compare the
graphs constructed from the real and the generated images to test the correct
placement of objects, without requiring the model to draw the objects in the same
exact locations (which would have defied its generative nature).

The \textbf{object detector and localizer} is based on the Inception-v3
architecture. We modify the last layer for object detection and replace it with
two heads. The first head is a linear layer with a sigmoid activation function to
serve as the object detector. It is trained with a binary cross-entropy loss. The
second head is a linear layer where we regress all the objects' coordinates.
This head is trained with an $L_2$-loss with a mask applied to only compute loss over
objects that occur in the ground truth image provided in the dataset. We
initialize the model using pre-trained weights trained over the ILSVRC12
(ImageNet) dataset and fine-tune on the \gls{codraw} or \gls{i-clevr} datasets.
Its performance is reported in the appendix.

\vspace*{1mm}
\noindent \textbf{Relational Similarity:}
To compare the arrangement of objects qualitatively, we use the above object
detector/localizer to determine the type and position of objects in the ground
truth and the generated image. We estimate a scene graph for each image, in which
the detected objects and the image center are the vertices. The directed edges
are given by the left-right and front-back relations between the vertices. To
compute a relational similarity metric on scene graphs, we determine how many
of the ground truth relations are present in the generated image:
\vspace*{-2mm}
\begin{align}
\RelSim(\EGgt,\EGgen) = \text{recall} \times \frac{|\EGgen \cap \EGgt|}{|\EGgt|}
\end{align}
\vspace*{-4mm}\\
where ``recall'' is the recall over objects detected in the generated image w.r.t
objects detected in the ground truth image. \EGgt{} is the set of
relational edges for the ground truth image corresponding to vertices common to
both ground truth images and generated images, and \EGgen{} is the set of
relational edges for the generated image corresponding to vertices common to
both ground truth images and generated images. The graph similarity for the
complete dataset is reported by taking the mean of the final time-step value for each example over the entire dataset. This metric is a lower bound on the actual relational accuracy,
as it penalizes relations based on how the objects are positioned in the
ground truth image. The same instructions may, however, permit different relationship graphs.
We present some examples of low-scoring to high-scoring images on this metric as well as additional discussion on $\RelSim$ in the appendix.

\subsection{Ablation Study}

\begin{table}[t]
	\small
	\centering
	\begin{tabu}to\linewidth{@{}X[1.8,l]*3{X[c,1]}X[c,1]X[c,1.2]@{}}
		\toprule
		& &&& \multicolumn{2}{c}{$D$ Fusion } \\\cmidrule{5-6}
		Model & $L_{D_{\text{wrong}}}$ & $f_{G_{t-1}}$ & $L_{\text{aux}}$ & concat & subtract\\
		\midrule
		Baseline &\xmark & \xmark & \xmark & \xmark & \xmark  \\
		Mismatch & \cmark & \xmark & \xmark & \xmark & \xmark \\
		$G$ prior  & \cmark & \cmark & \xmark & \xmark & \xmark   \\
		Aux & \cmark & \cmark & \cmark & \xmark & \xmark   \\
		$D$ Concat  & \cmark & \cmark & \cmark & \cmark & \xmark  \\
		$D$ Subtract   &\cmark & \cmark & \cmark & \xmark & \cmark \\ \bottomrule
   \end{tabu}
   \caption{Description of the components present in each model we test in the ablation study.}
   \label{table:legend}
   \vspace*{-3mm}
\end{table}

We experimented with different variations of our architecture to test the effect of each component.  We define the different instantiations of our architecture as follows:

\begin{itemize}[noitemsep,nolistsep]
\item{ \textbf{Baseline} The simplest version of our model. The discriminator loss only includes the adversarial terms $L_{\text{fake}}$ and $L_{\text{real}}$. The generator is only conditioned using the context-aware condition: $\widetilde{x}_{t} = G(z, h_{t})$. As for the discriminator, it has no access to the previous time-step image features. Only $\tilde{x}_{t}$ is encoded using $E_{D}$ and then passed to the discriminator $D$ without any fusion operations.}

\item{\textbf{Mismatch} The $L_{\text{wrong}}$ term is added to the discriminator loss. The rest of the model is similar to the baseline.}

\item{\textbf{\textit{G} prior} We condition the generator on the context-free condition $f_{G_{t-1}}$ in addition to $h_{t}$ as in equation~\eqref{eq:gen}}.

\item \textbf{Aux} In this model we add the $L_{\text{aux}}$ term to both the generator and discriminator losses. The loss functions for this model follow equations~\eqref{eq:D} and \eqref{eq:G}.

\item \textbf{\textit{D} Concat} In this model, the discriminator's input is the fused features from $x_{t-1}$ and ${x}_{t}$ (or $\tilde{x}_t$) encoded using $E_{D}$. The fusion is a simple concatenation across the channels dimension.

\item \textbf{\textit{D} Subtract} This is the same as ``\textit{D} Concat'' except for the fusion operation, which is an element-wise subtraction between the feature maps.

\item{ \textbf{Non-iterative} The non-iterative baseline uses the same model as the ``Mismatch'' baseline. All the input instructions are concatenated into one instruction and the final image is generated in a single-step.}
\end{itemize}

A summary of the components that are present for each model we test in the ablation study is provided in Table~\ref{table:legend}.

\subsection{Results}
\begin{figure*}
	\small
	\begin{tabu} to \textwidth {@{}XXXXX@{}}
		\centering \includegraphics[scale=0.5]{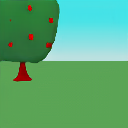} &
		\centering \includegraphics[scale=0.5]{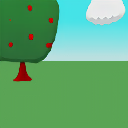} &
		\centering \includegraphics[scale=0.5]{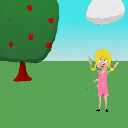} &
		\centering \includegraphics[scale=0.5]{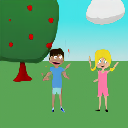} &
		\centering \includegraphics[scale=0.5]{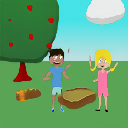}\\

		\dlgturn{Drawer}{ready}\par
		\dlgturn{Teller}{large apple tree left side trunk start 2 3 way up green and about 1 and 1 4 inches from left side}
		&
		\dlgturn{Teller}{big cloud right side almost touching apple tree 1 2 inch up into blue}
		&
		\dlgturn{Teller}{below the cloud full size girl her head touches top of green hands over head centered under cloud}
		&
		\dlgturn{Teller}{boy 1 1 2 to left of girl under right side of tree same height as girl facing right hands out to right holding a football in both}
		&
		\dlgturn{Teller}{sandbox medium size lower left corner facing right left side off screen lower right corner is equal to end of tree trunk}\par
		\dlgturn{Drawer}{yes}
	\end{tabu}

	\small
	\begin{tabu} to \textwidth {@{}X[0.8]X[0.8]X[0.9]X[0.9]X@{}}
		\centering \includegraphics[scale=0.5]{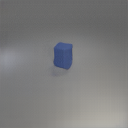} &
		\centering \includegraphics[scale=0.5]{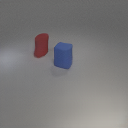} &
		\centering \includegraphics[scale=0.5]{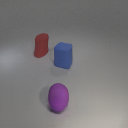} &
		\centering \includegraphics[scale=0.5]{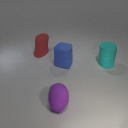} &
		\centering \includegraphics[scale=0.5]{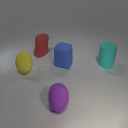}\\

		\iclvrex{Add a blue cube at the \par center}
		&

		\iclvrex{Add a red cylinder behind it on the left}
		&

		\iclvrex{Add a purple sphere in front of it on the right and in front of the blue cube}
		&

		\iclvrex{Add a cyan cylinder behind it on the right and in front of the red cylinder on the right}
		&

		\iclvrex{Add a yellow sphere in front of the red cylinder on the left and on the left of the blue cube}
	\end{tabu}
	\caption{Example images generated by our model ($D$ Subtract) on CoDraw (top row) and i-CLEVR (bottom row); shown with the provided instructions. We scale images from both datasets to 128x128 in a pre-processing step.}
   \label{fig:final-images}
   \vspace*{-3mm}
\end{figure*}
\noindent \textbf{Quantitative Results:}
 We present the results of the ablation study in Table~\ref{table:ablation}. As expected, among the iterative models, the \textit{Baseline} model has the weakest performance on all the metrics for both datasets. This is due to the fact that it needs to construct a completely new image from scratch every time-step; there is no consistency enforced between successive generations. As for the \textit{Mismatch} model, despite suffering from the same problems as the \textit{Baseline}, training $D$ to differentiate between wrong and right (image, instruction) pairs leads to generated images that better match the instructions. This is clear in Table~\ref{table:ablation} as the performance improves on all metrics compared to the \textit{Baseline}.

The \textit{$G$ prior} model tries to enforce consistency between generations by using the context-free condition $f_{G_{t-1}}$. Adding this condition leads to a significant improvement to all the metrics for the i-CLEVR dataset. However, for the CoDraw dataset, it shows a less significant improvement to recall and relational similarity, while precision degrades. These results can be explained by the fact that i-CLEVR has much more complex relationships between objects and the instructions have a strong dependence on the existing objects in the scene. Therefore, the model benefits from having access to how the objects were placed in the most recent iteration. As for CoDraw, the relationships among objects are relatively simpler. Nevertheless, adding the context-aware condition helps with placing the objects correctly as shown by the improvement in the relational similarity metric. A possible drawback from using the context-free condition is that it is harder to recover from past mistakes, especially if it has to do with a large objects. This drawback can explain the drop in precision.

For the \textit{Aux} model, it had different effects on the two datasets. For CoDraw, it helped improve recall and relational similarity, but caused a significant decrease in precision. For i-CLEVR, it helped improve precision with hurting the recall and relational similarity. This different behavior for each dataset can be explained by the types of objects that are present. While for CoDraw, there are objects that are almost always present like the girl or the boy, for i-CLEVR there is high randomness in objects presence. Adding the auxiliary objective encourages the model to make sure the frequent objects are present, leading to the increase in recall while hurting precision. Finally, we observe that giving $D$ access to the previous image $x_{t-1}$ shows improvement on almost all the metrics for both datasets. We also observe that subtraction fusion consistently performs better than concatenation fusion and outperforms all other models for both datasets. This indicates that encouraging the discriminator to focus on the modifications gives a better training signal for the generator.

The \textit{Non-iterative} model performs worse than all of the iterative models. This is likely because the language encoder has difficulty understanding dependencies and object relationships in a lengthy concatenated instruction. The benefit of using an iterative model is more visible in the i-CLEVR dataset since in it, the spatial relationships are always defined in terms of existing objects. This makes it very difficult to comprehend all the relationships across different turns in a single step. By having multiple steps, iterative generation makes this task easier. The results of this experiment make a case for iterative generation in complex text-conditional image generation tasks that have traditionally been performed non-iteratively.

\vspace*{1mm}
\noindent \textbf{Qualitative Results:}
 We present some examples of images generated by our model in
Figure~\ref{fig:final-images}. Due to space constraints, more example images
are provided in the appendix.
On \gls{codraw}, we observe
that the model is able to generate scenes consistent with the conversation and
generation history and gets most of the coarse details correct, such as large
objects and their relative positions. But it has difficulty in capturing fine-grained
details, such as tiny objects, facial expressions, and object poses. The model
also struggles when a single instruction asks to add several objects at once.
For \gls{i-clevr}, the model captures spatial relationships and colors very
accurately as demonstrated in Figure~\ref{fig:final-images}. However, in some instances, the model fails to add the fifth object when the image is
already crowded and there is no space left to add it without moving the others.
We also experimented with using an intermediate ground truth image as the initial image at test time and the model was able to generalize and place objects correctly in that scenario as well. The results of this experiment are presented in the appendix.

\vspace*{-1mm}
\section{Conclusion and Future Work}
\label{sec:conclusion}
We presented a recurrent \gls{gan} model for the \gls{geneva} task and show
that the model is able to draw reasonable images for the provided instructions iteratively. It also significantly outperforms the non-iterative baseline.
We presented an ablation study to highlight the contribution of different
components. Since this task can have several plausible solutions and no existing metric
can capture all of them, we proposed a relational similarity metric to capture the possible relationships. For future research directions, having a system that can also ask
questions from the user when it needs clarifications would potentially be even
more useful.
Collecting photo-realistic images, transitions between such images, and annotations in the form of instructions for these transitions is prohibitively expensive; hence, no photo-realistic dataset appropriate for this task publicly exists. Such datasets are needed to scale this task to photo-realistic images.

\section*{Acknowledgements}
\noindent We thank Philip Bachman for valuable discussions.

{
\fontsize{9.7pt}{12pt}\selectfont
\setlength{\bibsep}{1pt plus 0.3ex}
\bibliographystyle{IEEEtranN_fullname}
\bibliography{tell_draw_repeat}}

\clearpage
\twocolumn[{%
 \centering
 \bf
 \vspace*{3em}
 \Large{Tell, Draw, and Repeat: Generating and Modifying Images\\[0em]Based on Continual Linguistic Instruction}\\[1.5em]
 \Large Appendix\\[2.5em]
}]
\appendix
\input{supplementary.tex}

\end{document}

%% file: supplementary.tex

\section{Object detector and localizer network}
All of the evaluation metrics for the \gls{geneva} task rely on the object detector and localizer network and hence, it needs to have high detection and localization performance. We report the performance of the trained object detector and localizer network on the test set images of both \gls{codraw} and \gls{i-clevr} datasets in Table~\ref{table:obj-det-loc-perf}. 
\begin{table}[h]
	\small
	\centering
	\begin{tabu}to\linewidth{@{}X[l]*4{X[r]}@{}} \toprule
        Dataset & Precision$\uparrow$ & Recall$\uparrow$ & F1-Score$\uparrow$ & NRMSE$\downarrow$ \\ \midrule
        CoDraw  & 0.962     & 0.972  & 0.964    & 0.121 \\ 
        i-CLEVR & 1.000     & 1.000  & 1.000    & 0.060 \\ \bottomrule
    \end{tabu}
    \caption{Mean test set Precision, Recall, and F1-Score for the object detector and localizer network. Normalized Root Mean Squared Error (NRMSE) is the root mean square distance between the localizer's predicted and ground truth object centroids normalized by the image dimensions. $\uparrow$: higher is better, $\downarrow$: lower is better.}
    \label{table:obj-det-loc-perf}
\end{table}

\section{Relational Similarity metric: $\RelSim$}
\subsection{Additional details}
For both \gls{codraw} and \gls{i-clevr} datasets, we determine front-behind and left-right relationships by comparing the coordinates of their centre predicted by the object detector and localizer network. We run the network on both ground truth and generated images to predict the centre coordinates (rather than using perspective coordinates provided by the renderer  as these are only available for the ground truth for i-CLEVR).

\subsection{Appropriateness for evaluation}
\label{sec:appropriateness}
The \gls{codraw} and \gls{i-clevr} datasets are constructed such that there is at most one object of each object class per  image. Hence, we train the object detector to predict only binary presence of each object class and the localizer regresses  only one set of centroid coordinates per class. This design breaks if multiple instances of an object class are generated or if the object detector frequently misclassifies objects. However, qualitatively assessing the generated images, over-generation is rare and the object detector accuracy is very high (cf. Table~\ref{table:obj-det-loc-perf}).

Since all objects in ground truth scenes occur at most once, generations with multiple instances per class are out-of-distribution. The model cannot learn to exploit this flaw, since $\RelSim$ is not optimized during training. Thus, over-generation is not a failure mode we have observed. Additionally, $\RelSim$ is position-sensitive: over-generation would not necessarily produce the correct relative positions of objects since the object localizer only localizes one instance per class. For datasets with multiple instances per class, the $\RelSim$ metric should be modified such that the denominator is the union of ground-truth and predicted detections, which will penalize over-generation.

\subsection{Shortcomings}
Quantitative measures for attributes like ``boy kicking'' are currently a missing piece. We share this shortcoming with all text-to-image \gls{gan}-based methods and most of the conditional \gls{gan} literature. At the moment, conditional \glspl{gan}  are evaluated using \gls{is} and \gls{fid}, both of which do not account for attributes. An evaluation metric that accounts for attributes will be a valuable contribution for future research.

\subsection{Comparison with existing metrics}
The \gls{ssm} used by \citet{codraw} is well-suited for the setting of predicting object location and attributes. \gls{ssm} is a weighted score across recall and considers objects that face the wrong direction, incorrect expressions, poses, clip art size,  distance between object positions in ground truth and predicted image, and left-right and front-behind  relationships. \gls{ssm} achieves the highest score for exact reconstructions. In our case, we want to not just reward reconstructions but also plausible generations where left-right, front-behind relationships are correct. Our main focus here is to generate complete images instead of predicting object location and attributes. Several attributes, such as boy or girl poses / expressions, or object directions have lower detector accuracy and consequently would reduce metric reliability (cf. Section~\ref{sec:appropriateness}).

\subsection{Qualitative evaluation}
\begin{figure*}[!h]
   \scriptsize
   \setlength{\tabulinesep}{3pt}
    \begin{tabu} to \textwidth {X[0.2,m]X[0.2,m]X[1,m]}
      \centering
      \includegraphics[scale=0.5]{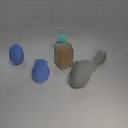} &
      \includegraphics[scale=0.5]{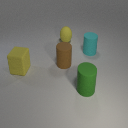} &
      {Objects detected in generated image: cube gray, cube brown, sphere gray, sphere blue\newline
      Objects detected in ground truth image: cube yellow, sphere yellow, cylinder green, cylinder brown, cylinder cyan\newline
      Recall: 0.00\newline
      $\RelSim$: 0.00\newline
      Explanation: None of the correct objects are drawn.}\\

      \centering
      \includegraphics[scale=0.5]{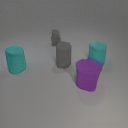} &
      \includegraphics[scale=0.5]{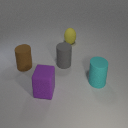} &
      {Objects detected in generated image: cube gray, cylinder gray, cylinder purple, cylinder cyan\newline
      Objects detected in ground truth image: cube purple, sphere yellow, cylinder gray, cylinder brown, cylinder cyan\newline
      Recall: 0.4\newline
      $\RelSim$: 0.25\newline
      Explanation: The cyan and gray cylinders are the only two objects detected in the generated image from the five ground truth objects detected.}\\

      \centering
      \includegraphics[scale=0.5]{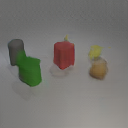} &
      \includegraphics[scale=0.5]{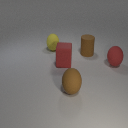} &
      {Objects detected in generated image: cube red, cube green, sphere brown, cylinder gray\newline
      Objects detected in ground truth image: cube red, sphere red, sphere brown, sphere yellow, cylinder brown\newline
      Recall: 0.4\newline
      $\RelSim$: 0.35\newline
      Explanation: The red cube and brown sphere are detected common to both images. Most of the relationships of these two and the center are correct.}\\

      \centering
      \includegraphics[scale=0.5]{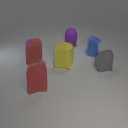} &
      \includegraphics[scale=0.5]{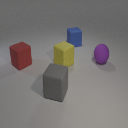} &
      {Objects detected in generated image: cube gray, cube red, cube yellow, sphere purple\newline
      Objects detected in ground truth image: cube gray, cube red, cube blue, cube yellow, sphere purple\newline
      Recall: 0.8\newline
      $\RelSim$: 0.45\newline
      Explanation: Only the blue cube is not detected in the generated image. Several spatial relationships of the common objects and the center are incorrect.}\\

      \centering
      \includegraphics[scale=0.5]{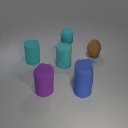} &
      \includegraphics[scale=0.5]{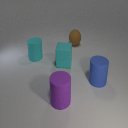} &
      {Objects detected in generated image: sphere brown, sphere cyan, cylinder blue, cylinder purple, cylinder cyan\newline
      Objects detected in ground truth image: cube cyan, sphere brown, cylinder blue, cylinder purple, cylinder cyan\newline
      Recall: 0.8\newline
      $\RelSim$: 0.675\newline
      Explanation: Cyan cube detected in ground truth image is missing from the generated image. Most spatial relationships of the common objects and center are correct.}\\

      \centering
      \includegraphics[scale=0.5]{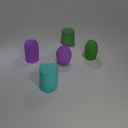} &
      \includegraphics[scale=0.5]{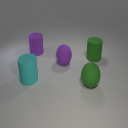} &
      {Objects detected in generated image: sphere green, sphere purple, cylinder green, cylinder purple, cylinder cyan\newline
      Objects detected in ground truth image: sphere green, sphere purple, cylinder green, cylinder purple, cylinder cyan\newline
      Recall: 1.0\newline
      $\RelSim$: 0.76\newline
      Explanation: All the objects are detected correctly but some of the spatial relationships are incorrect.}\\

      \centering
      \includegraphics[scale=0.5]{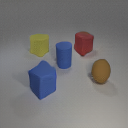} &
      \includegraphics[scale=0.5]{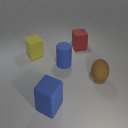} &
      {Objects detected in generated image: cube red, cube blue, cube yellow, sphere brown, cylinder blue\newline
      Objects detected in ground truth image: cube red, cube blue, cube yellow, sphere brown, cylinder blue\newline
      Recall: 1.00\newline
      $\RelSim$: 1.00\newline
      Explanation: All the objects are detected correctly and are in the correct relative positions.}\\
    \end{tabu}\vspace*{5mm}
 
    \caption{\textbf{Column 1:} Generated final image; \textbf{Column 2:} Ground truth final image; \textbf{Column 3:} List of objects detected in the generated and ground truth image, the recall on object detection, the value of the relational similarity ($\RelSim$) metric. The examples have been selected to qualitatively show examples with diverse score values between the minimum ($0$) and the maximum ($1$) values of the $\RelSim$ metric.}
    \label{fig:metric_high_low}
 \end{figure*}

We provide generated image examples with scores spread out between the minimum value ($0$) and maximum value ($1$) on the $\RelSim$ metric in Figure~\ref{fig:metric_high_low}. This is to provide readers with a more intuitive understanding of how the metric captures which spatial relationships match between the ground truth and the generated image.

\section{Generation Examples}
\begin{figure*}[!h]
	
	\scriptsize
	\begin{tabu} to \textwidth {XXXXX}
	\centering \includegraphics[scale=0.5]{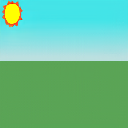}
	& \centering
	\includegraphics[scale=0.5]{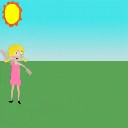} &
	\centering \includegraphics[scale=0.5]{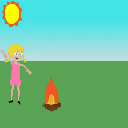}
	& \centering
	\includegraphics[scale=0.5]{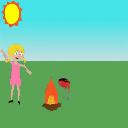}
	& \centering
	\includegraphics[scale=0.5]{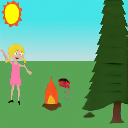}\\
	\centering
	\textbf{Drawer:} ready to draw ?
	\textbf{Teller:} Medium sun is on the left corner fully visible. 
	& 
	\centering
	\textbf{Teller:} Below sun sits a mad girl with legs on front she faces right and hand touches the left border a 1 2 head is above horizon. \textbf{Drawer:} ok.
	&
	\centering
	\textbf{Teller:} The girl is big. A fire is on front feet of girl like 1 2 ``. \textbf{Drawer:} ok.
	& 
	\centering
	\textbf{Teller:} A grill is just next to fire the grill is a little lower than top flame. \textbf{Drawer:} ok.
	&
	\centering
	\textbf{Teller:} A small pine is on right side 1 4 `` left side is cut also the tip is cut. \textbf{Drawer:} ok
\end{tabu}

	\scriptsize
	\begin{tabu} to \textwidth {XXXXX}
		\centering \includegraphics[scale=0.5]{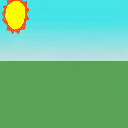}
		& \centering
		\includegraphics[scale=0.5]{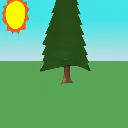} &
		\centering \includegraphics[scale=0.5]{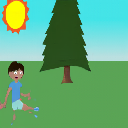}
		& \centering
		\includegraphics[scale=0.5]{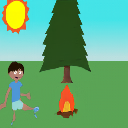}
		& \centering
		\includegraphics[scale=0.5]{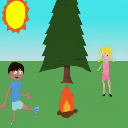}\\
		\centering
		\textbf{Teller:} In top left hand corner is medium sun cut off on top and on side. \textbf{Drawer:}  I am a patient worker ready to start.
		& 
		\centering
		\textbf{Teller:} In middle of screen is a medium pine tree trunk starts dead middle of screen. \textbf{Drawer:} Got it.
		&
		\centering
		\textbf{Teller:} A large boy is sitting cross legged almost in left corner slightly higher and to right he is facing right . \textbf{Drawer:} ok.
		&
		\centering
		\textbf{Teller:} Drink in right hand hot dog in left, to left of hot dog is a fire. \textbf{Drawer:} ok.
		& 
		\centering
		\textbf{Teller:} On right side is small girl angry face running her foot is cut off head just touches horizon. \textbf{Drawer:} ok.

	\end{tabu}

	\scriptsize
	\begin{tabu} to \textwidth {XXXXX}
	\centering \includegraphics[scale=0.5]{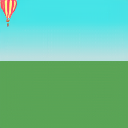}
	& \centering
	\includegraphics[scale=0.5]{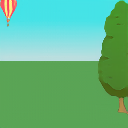} &
	\centering \includegraphics[scale=0.5]{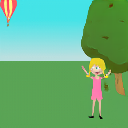}
	& \centering
	\includegraphics[scale=0.5]{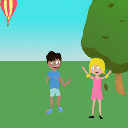}
	& \centering
	\includegraphics[scale=0.5]{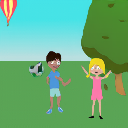}\\
	\centering
	\textbf{Teller:} Small hot air balloon in upper left corner touching left edge very top of balloon off top edge. \textbf{Drawer:} ok.
	& 
	\centering
	\textbf{Teller:} Med tree with hole in trunk at right side about 1 3 of it below horizon and right edge of it off screen. \textbf{Drawer:} ok.
	&
	\centering
	\textbf{Teller:} Big crying girl sitting on ground legs outstretched facing left top of her head touching bottom left corner of tree trunk. \textbf{Drawer:} ok.
	& 
	\centering
	\textbf{Teller:} Big standing boy arms in air facing right toes touching bottom edge of page slightly left of center. \textbf{Drawer:} ok.
	&
	\centering
	\textbf{Teller:} Soccer ball on ground between boy and girl about level with boy 's hips
	 . \textbf{Drawer:} ok.
\end{tabu}
	
	\scriptsize
	\begin{tabu} to \textwidth {XXXXX}
		\centering \includegraphics[scale=0.5]{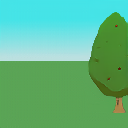}
		& \centering
		\includegraphics[scale=0.5]{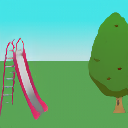} &
		\centering \includegraphics[scale=0.5]{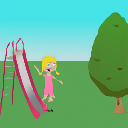}
		& \centering
		\includegraphics[scale=0.5]{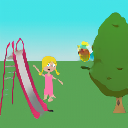}
		& \centering
		\includegraphics[scale=0.5]{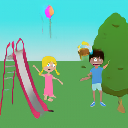}\\
		\centering
		\textbf{Drawer:} Ready.
		\textbf{Teller:} Medium bushy tree right side top 2 3 above horizon right side touching right side of tree. 
		& 
		\centering
		\textbf{Teller:} Left side medium slide facing right half ladder cut off at left one step is in sky. \textbf{Drawer:} Done.
		&
		\centering
		\textbf{Teller:} head covering bottom of slide part bum by bottom sitting legs out happy girl facing right. \textbf{Drawer:}  Done.
		& 
		\centering
		\textbf{Teller:} By her back hand is a medium beach ball her hand is touching the right side of it from hand holding medium party ballo. \textbf{Drawer:} Done.
		&
		\centering
		\textbf{Teller:} Actual balloons part is over horizon back of head touching left side of tree running happy boy facing girl wearing rainbow hat. \textbf{Drawer:} Done.
	\end{tabu}

	\scriptsize
	\begin{tabu} to \textwidth {XXXXX}
		\centering \includegraphics[scale=0.5]{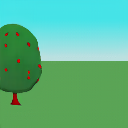}
		& \centering
		\includegraphics[scale=0.5]{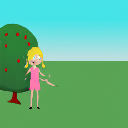} &
		\centering \includegraphics[scale=0.5]{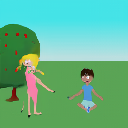}
		& \centering
		\includegraphics[scale=0.5]{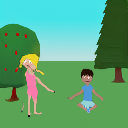}
		& \centering
		\includegraphics[scale=0.5]{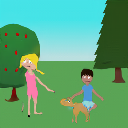}\\
		\centering
		\textbf{Drawer:} Ready.
		\textbf{Teller:} On the left is a small apple tree slightly cut at the left edge.
		\textbf{Drawer:} Done.
		& 
		\centering
		\textbf{Teller:} In front of the tree facing right stands a happy girl waving \textbf{Drawer:} Done.
		&
		\centering
		\textbf{Teller:} In the middle of the field standing facing right his hands to the right. \textbf{Drawer:} Who is standing in the middle and what expression?
		& 
		\centering
		\textbf{Teller:} Is a boy he is angry. \textbf{Drawer:} Done.
		\textbf{Teller:} On the right is a medium pine tree slightly cut at the right and at the top. \textbf{Drawer:} Done.
		&
		\centering
		\textbf{Teller:} In front of the tree is a dog facing left. \textbf{Drawer:} Done.
	\end{tabu}
	
	\scriptsize
	\begin{tabu} to \textwidth {XXXXX}
		\centering \includegraphics[scale=0.5]{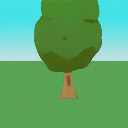}
		& \centering
		\includegraphics[scale=0.5]{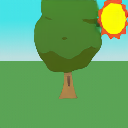} &
		\centering \includegraphics[scale=0.5]{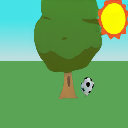}
		& \centering
		\includegraphics[scale=0.5]{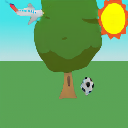}
		& \centering
		\includegraphics[scale=0.5]{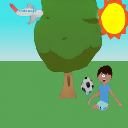}\\
		\centering
		\textbf{Drawer:} Start.
		\textbf{Teller:} Bushy tree in middle.
		\textbf{Drawer:} Ok.
		& 
		\centering
		\textbf{Teller:} Full sun right corner. \textbf{Drawer:} Yes.
		&
		\centering
		\textbf{Teller:} Soccer ball bottom right corner. \textbf{Drawer:} Done.
		& 
		\centering
		\textbf{Teller:} Plane top left corner . \textbf{Drawer:} Ok.
		&
		\centering
		\textbf{Teller:} Boy between sun and soccer ball. \textbf{Drawer:} Boy direction.
		\textbf{Teller:} Hands out facing left. \textbf{Drawer:} Ok.
	\end{tabu}\vspace*{5mm}
	
	\caption{Generation examples from our best model ($D$ Subtract) for the \gls{codraw} dataset.}
	\label{fig:codraw_examples}
\end{figure*}

\begin{figure*}[!hb]
	\scriptsize
	\begin{tabu} to \textwidth {XXXXX}
		\centering \includegraphics[scale=0.5]{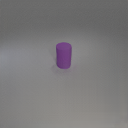}
		& \centering
		\includegraphics[scale=0.5]{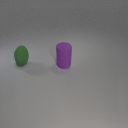} &
		\centering \includegraphics[scale=0.5]{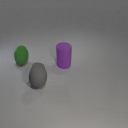}
		& \centering
		\includegraphics[scale=0.5]{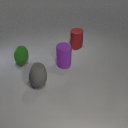}
		& \centering
		\includegraphics[scale=0.5]{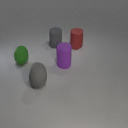}\\
		\centering
		Add a purple cylinder at the center
		& 
		\centering
		Add a green sphere on the left of it
		&
		\centering
		Add a gray sphere in front of it on the right and in front of the purple cylinder on the left
		& 
		\centering
		Add a red cylinder behind it on the right and behind the green sphere on the right
		&
		\centering
		Add a gray cylinder behind the green sphere on the right and behind the purple cylinder on the left
	\end{tabu}

\scriptsize
\begin{tabu} to \textwidth {XXXXX}
	\centering \includegraphics[scale=0.5]{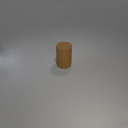}
	& \centering
	\includegraphics[scale=0.5]{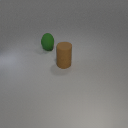} &
	\centering \includegraphics[scale=0.5]{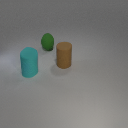}
	& \centering
	\includegraphics[scale=0.5]{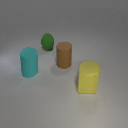}
	& \centering
	\includegraphics[scale=0.5]{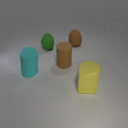}\\
	\centering
	Add a brown cylinder at the center
	& 
	\centering
	Add a green sphere behind it on the left
	&
	\centering
	Add a cyan cylinder in front of it on the left and in front of the brown cylinder on the left
	& 
	\centering
	Add a yellow cube in front of the green sphere on the right and in front of the brown cylinder on the right
	&
	\centering
	Add a brown sphere behind it and behind the cyan cylinder on the right
\end{tabu}
	
	\scriptsize
	\begin{tabu} to \textwidth {XXXXX}
		\centering \includegraphics[scale=0.5]{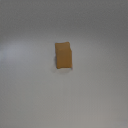}
		& \centering
		\includegraphics[scale=0.5]{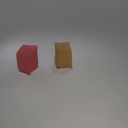} &
		\centering \includegraphics[scale=0.5]{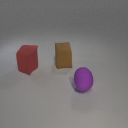}
		& \centering
		\includegraphics[scale=0.5]{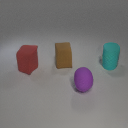}
		& \centering
		\includegraphics[scale=0.5]{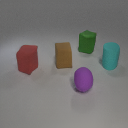}\\
		\centering
		Add a brown cube at the center
		& 
		\centering
		Add a red cube in front of it on the left
		&
		\centering
		Add a purple sphere in front of it on the right and in front of the brown cube on the right
		& 
		\centering
		Add a cyan cylinder behind it on the right and on the right of the brown cube
		&
		\centering
		Add a green cube behind the red cube on the right and behind the brown cube on the right
	\end{tabu}
	
	\scriptsize
	\begin{tabu} to \textwidth {XXXXX}
		\centering \includegraphics[scale=0.5]{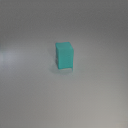}
		& \centering
		\includegraphics[scale=0.5]{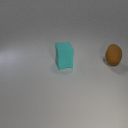} &
		\centering \includegraphics[scale=0.5]{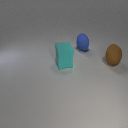}
		& \centering
		\includegraphics[scale=0.5]{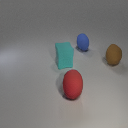}
		& \centering
		\includegraphics[scale=0.5]{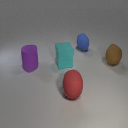}\\
		\centering
		Add a cyan cube at the center
		& 
		\centering
		Add a brown sphere on the right of it
		&
		\centering
		Add a blue sphere behind it on the left and behind the cyan cube on the right
		& 
		\centering
		Add a red sphere in front of the brown sphere on the left and in front of the cyan cube on the right
		&
		\centering
		Add a purple cylinder behind it on the left and in front of the cyan cube on the left
	\end{tabu}

	\scriptsize
	\begin{tabu} to \textwidth {XXXXX}
		\centering \includegraphics[scale=0.5]{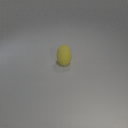}
		& \centering
		\includegraphics[scale=0.5]{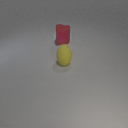} &
		\centering \includegraphics[scale=0.5]{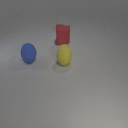}
		& \centering
		\includegraphics[scale=0.5]{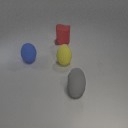}
		& \centering
		\includegraphics[scale=0.5]{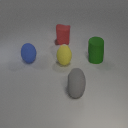}\\
		\centering
		Add a yellow sphere at the center
		& 
		\centering
		Add a red cube behind it
		&
		\centering
		Add a blue sphere in front of it on the left and behind the yellow sphere on the left
		& 
		\centering
		Add a gray sphere in front of it on the right and in front of the red cube on the right
		&
		\centering
		\textbf{Add a green cylinder in front of the blue sphere on the right and behind the yellow sphere on the right}
	\end{tabu}
	
	\scriptsize
	\begin{tabu} to \textwidth {XXXXX}
		\centering \includegraphics[scale=0.5]{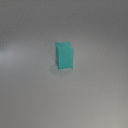}
		& \centering
		\includegraphics[scale=0.5]{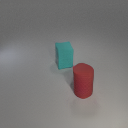} &
		\centering \includegraphics[scale=0.5]{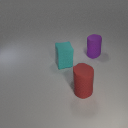}
		& \centering
		\includegraphics[scale=0.5]{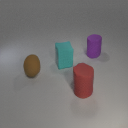}
		& \centering
		\includegraphics[scale=0.5]{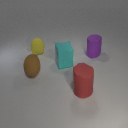}\\
		\centering
		Add a cyan cube at the center
		& 
		\centering
		Add a red cylinder in front of it on the right
		&
		\centering
		Add a purple cylinder behind it on the right and behind the cyan cube on the right
		& 
		\centering
		\textbf{Add a brown sphere in front of it on the left and in front of the red cylinder on the left}
		&
		\centering
		Add a yellow sphere in front of the purple cylinder on the left and behind the red cylinder on the left
	\end{tabu}\vspace*{5mm}
	
	\caption{Generation examples from our best model ($D$ Subtract) for the \gls{i-clevr} dataset. Instructions where the model made a mistake are marked in bold.}
	\label{fig:clevr_examples}
\end{figure*}

\begin{figure*}[!h]
   \scriptsize
   \begin{tabu} to \textwidth {XXXXX}
       \centering \includegraphics[scale=0.5]{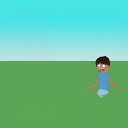} &
       \centering \includegraphics[scale=0.5]{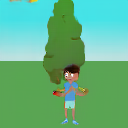} &
       \centering \includegraphics[scale=0.5]{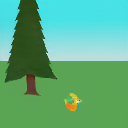} &
       \centering \includegraphics[scale=0.5]{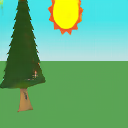} &
       \centering \includegraphics[scale=0.5]{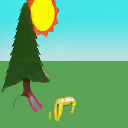} \\
       \centering \textbf{Teller:} small pine on right most of tree cut off on right and top big shocked running mike on right facing left his elbow on the left edge of tree &
       \centering \textbf{Teller:} big sun in middle 1 2 cut off on top small oak tree on left hole facing right 1 4 from top &
       \centering \textbf{Teller:} big slide on left facing right slide is in front of tree &
       \centering \textbf{Teller:} big angry sitting with legs out jenny is on ground in front of end of slide &
       \centering \textbf{Teller:} big cat is under slide mike 's head is over horizon jenny is 1 2 below horizon 
   \end{tabu}

   \scriptsize
   \begin{tabu} to \textwidth {XXXXX}
       \centering \includegraphics[scale=0.5]{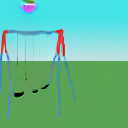} &
       \centering \includegraphics[scale=0.5]{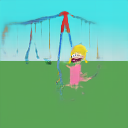} &
       \centering \includegraphics[scale=0.5]{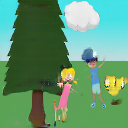} &
       \centering \includegraphics[scale=0.5]{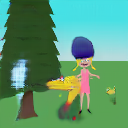} &
       \centering \includegraphics[scale=0.5]{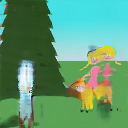} \\
       \centering \textbf{Drawer:} ready when you are describe away \textbf{Teller:} large swing on the left facing right &
       \centering \textbf{Teller:} looks like a happy girl standing on the 2nd swing facing right no teeth &
       \centering \textbf{Teller:} sad facing right feet almost at the bottom an inch from the left \textbf{Drawer:} are his arms out or left one up &
       \centering \textbf{Teller:} one arm in the air \textbf{Drawer:} got it \textbf{Teller:} there is a sun on the top right a little bit of it is seen it 's behind the tree &
       \centering \textbf{Teller:} under the tree is a bee facing left \textbf{Drawer:} got it 
   \end{tabu}

   \scriptsize
   \begin{tabu} to \textwidth {XXXXX}
       \centering \includegraphics[scale=0.5]{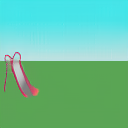} &
       \centering \includegraphics[scale=0.5]{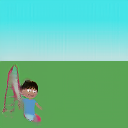} &
       \centering \includegraphics[scale=0.5]{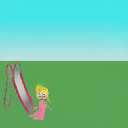} &
       \centering \includegraphics[scale=0.5]{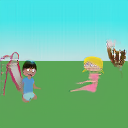} &
       \centering \includegraphics[scale=0.5]{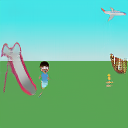} \\
       \centering \textbf{Teller:} there is a small slide on the left side facing right tip of the top of the slide is above horizon &
       \centering \textbf{Teller:} the boy with sad face is sliding down two legs kicked out there 's a baseball near the end of the slide \textbf{Drawer:} ready &
       \centering \textbf{Teller:} the boy is probably medium size there is also a medium girl on the right edge smiling and jumping &
       \centering \textbf{Teller:} she is facing left and the top of her head is touching the horizon she has a baseball glove on her left hand &
       \centering \textbf{Teller:} there is an airplane facing right at the top right of the picture let me know when you 're ready for me to check 
   \end{tabu}

   \scriptsize
   \begin{tabu} to \textwidth {XXXXX}
       \centering \includegraphics[scale=0.5]{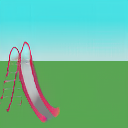} &
       \centering \includegraphics[scale=0.5]{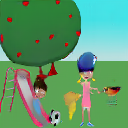} &
       \centering \includegraphics[scale=0.5]{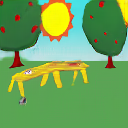} &
       \centering \includegraphics[scale=0.5]{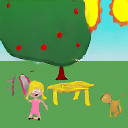} &
       \centering \includegraphics[scale=0.5]{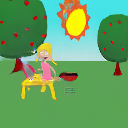} \\
       \centering \textbf{Drawer:} ready \textbf{Teller:} medium slide on left of screen facing right about 1 2 `` from left surprised boy sitting at bottom of slide &
       \centering \textbf{Teller:} small tree in back of slide to right a bit and then an apple tree to the right of that tree about 1 2 `` apart &
       \centering \textbf{Teller:} sun between the trees bottom left of sun blocked by tree on left \textbf{Drawer:} got it &
       \centering \textbf{Teller:} surprised girl on right of screen about 1 2 inch from right border hands in the air facing left \textbf{Drawer:} got it &
       \centering \textbf{Teller:} tennis ball beneath her left foot \textbf{Drawer:} got it 
   \end{tabu}

   \scriptsize
   \begin{tabu} to \textwidth {XXXXX}
       \centering \includegraphics[scale=0.5]{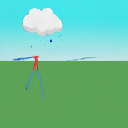} &
       \centering \includegraphics[scale=0.5]{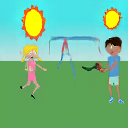} &
       \centering \includegraphics[scale=0.5]{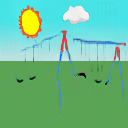} &
       \centering \includegraphics[scale=0.5]{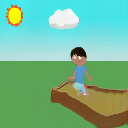} &
       \centering \includegraphics[scale=0.5]{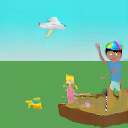} \\
       \centering \textbf{Teller:} left side big swing set left behind leg cut off edge and left corner small sun covered partially by small plain cloud &
       \centering \textbf{Teller:} right side medium sand basket dune facing left in front of dune sad cross leg sitting crying girl facing right \textbf{Drawer:} ready &
       \centering \textbf{Teller:} next to girl cat sitting looking at her her hand is hidden by cat 's head &
       \centering \textbf{Teller:} in front of the sandbox near the girl in the corner boy sad cross leg looking right &
       \centering \textbf{Teller:} behind boy and behind girl 's hand spring bee body hidden by boy facing right and i will check 
   \end{tabu}\vspace*{5mm}

   \caption{Random selection of examples generated by our Baseline model for the CoDraw dataset.}
   \label{fig:random_examples_baseline}
\end{figure*}

\begin{figure*}[!h]
   \scriptsize
   \begin{tabu} to \textwidth {XXXXX}
       \centering \includegraphics[scale=0.5]{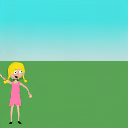} &
       \centering \includegraphics[scale=0.5]{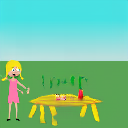} &
       \centering \includegraphics[scale=0.5]{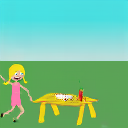} &
       \centering \includegraphics[scale=0.5]{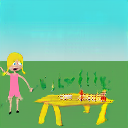} &
       \centering \includegraphics[scale=0.5]{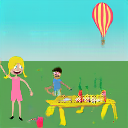} \\
       \centering \textbf{Drawer:} ready \textbf{Teller:} 1 girl happy running facing right 0 2 inch from bottom to top and 0 2 inches from left to right with a chef &
       \centering \textbf{Teller:} hat on her a table 1 and a half inches from left to right 1 2 inches from bottom to top with a pizza in the middle &
       \centering \textbf{Teller:} and and a hot dog facing left to the right of the pizza a fire 0 1 inches from bottom to top 0 4 inches from right to left and above a &
       \centering \textbf{Teller:} tent facing left and top of the tent is above the horizon line 0 1 inches and right is cover a little bit and above &
       \centering \textbf{Teller:} a air balloon small like 1 2 inches from right to left and 0 2 inches from top to bottom and that 's it 
   \end{tabu}

   \scriptsize
   \begin{tabu} to \textwidth {XXXXX}
       \centering \includegraphics[scale=0.5]{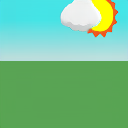} &
       \centering \includegraphics[scale=0.5]{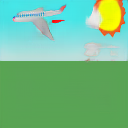} &
       \centering \includegraphics[scale=0.5]{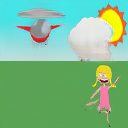} &
       \centering \includegraphics[scale=0.5]{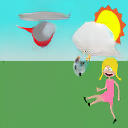} &
       \centering \includegraphics[scale=0.5]{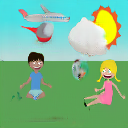} \\
       \centering \textbf{Drawer:} what 's in the picture \textbf{Teller:} in the top right is a sun covered partially by clouds \textbf{Drawer:} is it a large sun &
       \centering \textbf{Teller:} in the middle left is a helicopter \textbf{Drawer:} which way is it facing &
       \centering \textbf{Teller:} yes a large sun \textbf{Teller:} heli is facing to the right \textbf{Teller:} tail is to the left \textbf{Teller:} on bottom right is a girl in pink with left arm raised &
       \centering \textbf{Teller:} on his mouth is an o shape \textbf{Teller:} to the right of the boy is a dog with a blue collar &
       \centering \textbf{Teller:} above the boys left hand in the blue sky is a yellow ball 
   \end{tabu}

   \scriptsize
   \begin{tabu} to \textwidth {XXXXX}
       \centering \includegraphics[scale=0.5]{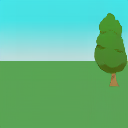} &
       \centering \includegraphics[scale=0.5]{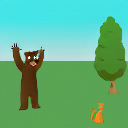} &
       \centering \includegraphics[scale=0.5]{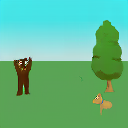} &
       \centering \includegraphics[scale=0.5]{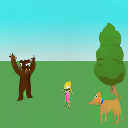} &
       \centering \includegraphics[scale=0.5]{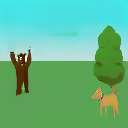} \\
       \centering \textbf{Drawer:} may you please tell the first thing to draw \textbf{Teller:} there is a small tree on the right side of the scene sort of in the background \textbf{Drawer:} &
       \centering \textbf{Teller:} there is a bear to the left but in the foreground of the tree \textbf{Drawer:} what next &
       \centering \textbf{Teller:} the bear is small \textbf{Drawer:} \textbf{Teller:} the girl is to the left facing left looking at the bear with her leg out scared facing right \textbf{Drawer:} &
       \centering \textbf{Teller:} there is a small dog below the girl and a angry boy to left facing left with a racket in the left hand \textbf{Drawer:} what is next &
       \centering \textbf{Teller:} there is a small helicopter at the top in the sky in between the boy and girl \textbf{Drawer:} right above the bear 
   \end{tabu}

   \scriptsize
   \begin{tabu} to \textwidth {XXXXX}
       \centering \includegraphics[scale=0.5]{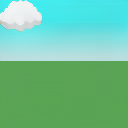} &
       \centering \includegraphics[scale=0.5]{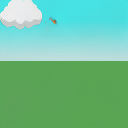} &
       \centering \includegraphics[scale=0.5]{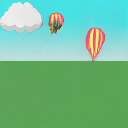} &
       \centering \includegraphics[scale=0.5]{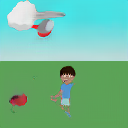} &
       \centering \includegraphics[scale=0.5]{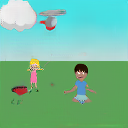} \\
       \centering \textbf{Teller:} there is a cloud in the top left corner it is cut off on the top and left sides two puffs on the right and 3 puffs on the bottom &
       \centering \textbf{Teller:} an inch from the right of the cloud are small balloons the orange balloon is on the right &
       \centering \textbf{Teller:} airplane 1 4 inch to the right of the balloon the nose of the plane is in line with the yellow balloon it faces left &
       \centering \textbf{Teller:} angry boy sitting below the plane facing left about 1 2 inch grass above and below him &
       \centering \textbf{Teller:} a girl sits to the left of the boy facing him feet almost touching surprised wearing viking hat top brown part of hat touches horizon 
   \end{tabu}

   \scriptsize
   \begin{tabu} to \textwidth {XXXXX}
       \centering \includegraphics[scale=0.5]{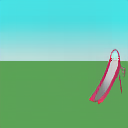} &
       \centering \includegraphics[scale=0.5]{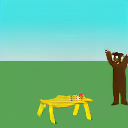} &
       \centering \includegraphics[scale=0.5]{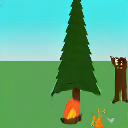} &
       \centering \includegraphics[scale=0.5]{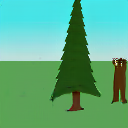} &
       \centering \includegraphics[scale=0.5]{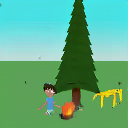} \\
       \centering \textbf{Drawer:} ready \textbf{Teller:} a small bear close to the right side one finger off picture small sliding on his left side bear left foot touching the sliding &
       \centering \textbf{Teller:} table medium in the center front &
       \centering \textbf{Teller:} medium pine tree behind the table one inch &
       \centering \textbf{Teller:} sad girl standing far left part hand missing sad face medium &
       \centering \textbf{Teller:} boy sitting on her right side look mad 
   \end{tabu}\vspace*{5mm}

   \caption{Random selection of examples generated by our Mismatch model for the CoDraw dataset.}
   \label{fig:random_examples_mismatch}
\end{figure*}

\begin{figure*}[!h]
   \scriptsize
   \begin{tabu} to \textwidth {XXXXX}
       \centering \includegraphics[scale=0.5]{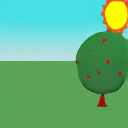} &
       \centering \includegraphics[scale=0.5]{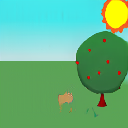} &
       \centering \includegraphics[scale=0.5]{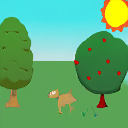} &
       \centering \includegraphics[scale=0.5]{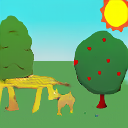} &
       \centering \includegraphics[scale=0.5]{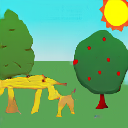} \\
       \centering \textbf{Teller:} medium sun on your right hand with a small half apple tree under it &
       \centering \textbf{Teller:} small snake left of apple tree snake facing left &
       \centering \textbf{Teller:} small bushy tree is on your left hand side hole facing left 1 2 inch from side 1 inch from top &
       \centering \textbf{Teller:} large table in left hand corner slanted right north with medium owl on right end facing right &
       \centering \textbf{Teller:} bushy tree looks like it 's sitting on table 
   \end{tabu}

   \scriptsize
   \begin{tabu} to \textwidth {XXXXX}
       \centering \includegraphics[scale=0.5]{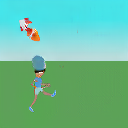} &
       \centering \includegraphics[scale=0.5]{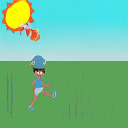} &
       \centering \includegraphics[scale=0.5]{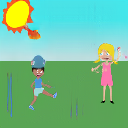} &
       \centering \includegraphics[scale=0.5]{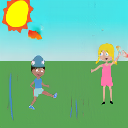} &
       \centering \includegraphics[scale=0.5]{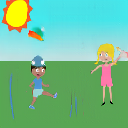} \\
       \centering \textbf{Drawer:} i 'm ready \textbf{Teller:} boy flying a kite &
       \centering \textbf{Teller:} wearing blue t shirt \textbf{Teller:} blue shoes \textbf{Teller:} lighter blue shorts \textbf{Drawer:} got it \textbf{Teller:} has black hair \textbf{Drawer:} i have all of him \textbf{Teller:} sunny outside &
       \centering \textbf{Teller:} a girl next to a grill &
       \centering \textbf{Teller:} i am sorry only one girl in the scene \textbf{Teller:} next to a girl \textbf{Teller:} on the grill \textbf{Drawer:} is a burger \textbf{Teller:} there are hamburgers &
       \centering \textbf{Teller:} 3 of them \textbf{Teller:} she is wearing pink overall holding ketchup in her hand and she is wearing glasses on 
   \end{tabu}

   \scriptsize
   \begin{tabu} to \textwidth {XXXXX}
       \centering \includegraphics[scale=0.5]{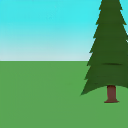} &
       \centering \includegraphics[scale=0.5]{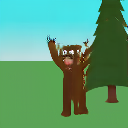} &
       \centering \includegraphics[scale=0.5]{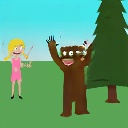} &
       \centering \includegraphics[scale=0.5]{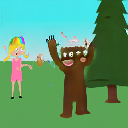} &
       \centering \includegraphics[scale=0.5]{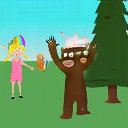} \\
       \centering \textbf{Drawer:} ready when you are \textbf{Teller:} right side medium pine tree cut in half by right edge and cut top \textbf{Drawer:} got it &
       \centering \textbf{Teller:} left from pine tree big size bear facing left head above horizon &
       \centering \textbf{Teller:} left side running angry big size girl facing right horn hat holding football in right hand our right &
       \centering \textbf{Teller:} now behind girl they overlap running angry boy facing right purple glasses witch hat on facing right half body above horizon \textbf{Drawer:} got it &
       \centering \textbf{Teller:} movie girl down a bit move boy right a bit lines with girl next to girls what is a frisbee 
   \end{tabu}

   \scriptsize
   \begin{tabu} to \textwidth {XXXXX}
       \centering \includegraphics[scale=0.5]{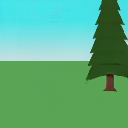} &
       \centering \includegraphics[scale=0.5]{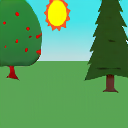} &
       \centering \includegraphics[scale=0.5]{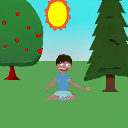} &
       \centering \includegraphics[scale=0.5]{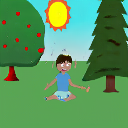} &
       \centering \includegraphics[scale=0.5]{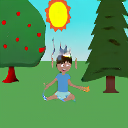} \\
       \centering \textbf{Drawer:} what is in the scenery of the image \textbf{Teller:} medium pine tree on right trunk is about 1 inch from bottom and trunk is 1 2 from right top is cut off on top and right \textbf{Drawer:} what 's next &
       \centering \textbf{Teller:} on left is medium apple tree cut off on left trunk is halfway down grass big sun in center of two trees \textbf{Drawer:} what 's next &
       \centering \textbf{Teller:} on the right side of apple tree trunk is sad boy sitting legs out facing right hes 1 inch from bottom very close to trunk \textbf{Drawer:} what 's next &
       \centering \textbf{Teller:} to right of boy align with his hand is a fire to right of fire is girl kneeling smiling one arm up girl and boy small \textbf{Drawer:} what 's next &
       \centering \textbf{Teller:} girls right hand overlaps pine tree a little left side of sun is overlapped by apple tree below fire is ketchup left mustard right \textbf{Drawer:} next 
   \end{tabu}

   \scriptsize
   \begin{tabu} to \textwidth {XXXXX}
       \centering \includegraphics[scale=0.5]{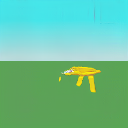} &
       \centering \includegraphics[scale=0.5]{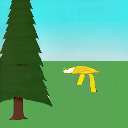} &
       \centering \includegraphics[scale=0.5]{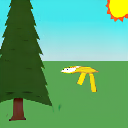} &
       \centering \includegraphics[scale=0.5]{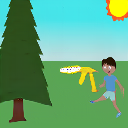} &
       \centering \includegraphics[scale=0.5]{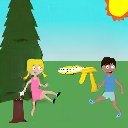} \\
       \centering \textbf{Teller:} ready \textbf{Drawer:} what what is the first object and location \textbf{Teller:} small table middle of green \textbf{Drawer:} next &
       \centering \textbf{Teller:} there is a pine to the left med size upper peck can not see \textbf{Drawer:} next &
       \centering \textbf{Teller:} med sun in upper right corner \textbf{Drawer:} any of it cut off &
       \centering \textbf{Teller:} no sun is whole straight down from sun is a boy standing with a laugh on face \textbf{Drawer:} facing left &
       \centering \textbf{Teller:} yes boy is almost at bottom of screen \textbf{Drawer:} what else \textbf{Teller:} there is a girl standing at corner of table looks likes she is running smiling \textbf{Drawer:} next 
   \end{tabu}\vspace*{5mm}

   \caption{Random selection of examples generated by our $G$ prior model for the CoDraw dataset.}
   \label{fig:random_examples_g-prior}
\end{figure*}

\begin{figure*}[!h]
   \scriptsize
   \begin{tabu} to \textwidth {XXXXX}
       \centering \includegraphics[scale=0.5]{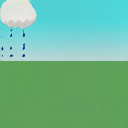} &
       \centering \includegraphics[scale=0.5]{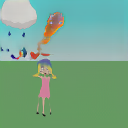} &
       \centering \includegraphics[scale=0.5]{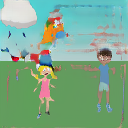} &
       \centering \includegraphics[scale=0.5]{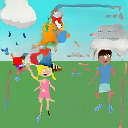} &
       \centering \includegraphics[scale=0.5]{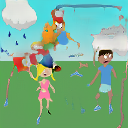} \\
       \centering \textbf{Drawer:} go \textbf{Teller:} large rain cloud left corner touches side cut off on top drops almost touch grass \textbf{Drawer:} next &
       \centering \textbf{Teller:} large rocket on right tip of cloud flying left with very small girl sad legs out sitting on its upper wing \textbf{Drawer:} sitting on rocket the rocket is middle scene &
       \centering \textbf{Teller:} rocket overcloud large regular cloud on right side cut off on top and side a bit surprised boy legs out facing left under cloud \textbf{Drawer:} next &
       \centering \textbf{Teller:} cat facing boy 1 2 inch to left of his feet \textbf{Drawer:} next &
       \centering \textbf{Teller:} i will check and send adjustments \textbf{Drawer:} yes i do n't have the girl tell me where is the girl \textbf{Teller:} she is sitting on the rockets upper wing her back arm is under the window \textbf{Drawer:} check 
   \end{tabu}

   \scriptsize
   \begin{tabu} to \textwidth {XXXXX}
       \centering \includegraphics[scale=0.5]{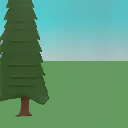} &
       \centering \includegraphics[scale=0.5]{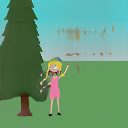} &
       \centering \includegraphics[scale=0.5]{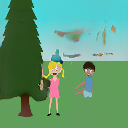} &
       \centering \includegraphics[scale=0.5]{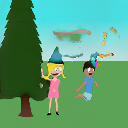} &
       \centering \includegraphics[scale=0.5]{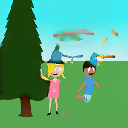} \\
       \centering \textbf{Drawer:} where is jenny and mike \textbf{Teller:} on left hand side of the screen 2 inches above bottom is a pine tree cut off at top &
       \centering \textbf{Teller:} straight down from tree is jenny legs crossed facing right right arm in the air &
       \centering \textbf{Teller:} next to jenny is mike same level standing facing right with arms out mouth open &
       \centering \textbf{Teller:} on the right hand side inch from the bottom is a duck facing jenny and mike &
       \centering \textbf{Teller:} straight above duck is soccer ball 
   \end{tabu}

   \scriptsize
   \begin{tabu} to \textwidth {XXXXX}
       \centering \includegraphics[scale=0.5]{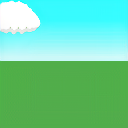} &
       \centering \includegraphics[scale=0.5]{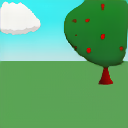} &
       \centering \includegraphics[scale=0.5]{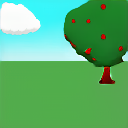} &
       \centering \includegraphics[scale=0.5]{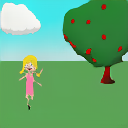} &
       \centering \includegraphics[scale=0.5]{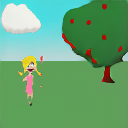} \\
       \centering \textbf{Teller:} large cloud in left corner top and left are off screen &
       \centering \textbf{Teller:} large apple tree right side top of trunk lines up with horizon right side and top of screen &
       \centering \textbf{Teller:} in front of trunk little over to the right of trunk by right side is a soccer ball &
       \centering \textbf{Teller:} left side large girl angry face sitting cross legged facing right &
       \centering \textbf{Teller:} neck on horizon line she 's holding a baseball in her up hand and wearing rainbow hat 
   \end{tabu}

   \scriptsize
   \begin{tabu} to \textwidth {XXXXX}
       \centering \includegraphics[scale=0.5]{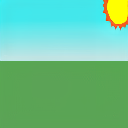} &
       \centering \includegraphics[scale=0.5]{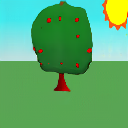} &
       \centering \includegraphics[scale=0.5]{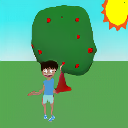} &
       \centering \includegraphics[scale=0.5]{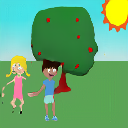} &
       \centering \includegraphics[scale=0.5]{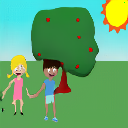} \\
       \centering \textbf{Drawer:} what do we have \textbf{Teller:} med sun right corner \textbf{Drawer:} and &
       \centering \textbf{Teller:} middle of green with trees half in blue half in green is a apple tree med size \textbf{Drawer:} and &
       \centering \textbf{Teller:} med boy standing on right of tree \textbf{Drawer:} face expression and where are his hands \textbf{Teller:} he has a tennis ball in right hand he is smiling showing teeth \textbf{Drawer:} and &
       \centering \textbf{Teller:} right arm sticking out across from him is a girl almost to the left edge left arm down right hand with a ball glove \textbf{Drawer:} smiling please give more elaborations &
       \centering \textbf{Teller:} yes smiling looking right she is standing her middle is on the line of green \textbf{Drawer:} and 
   \end{tabu}

   \scriptsize
   \begin{tabu} to \textwidth {XXXXX}
       \centering \includegraphics[scale=0.5]{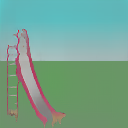} &
       \centering \includegraphics[scale=0.5]{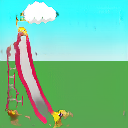} &
       \centering \includegraphics[scale=0.5]{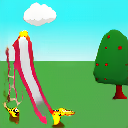} &
       \centering \includegraphics[scale=0.5]{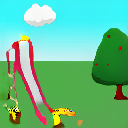} &
       \centering \includegraphics[scale=0.5]{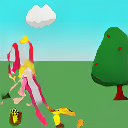} \\
       \centering \textbf{Teller:} large slide to the left facing right with owl sitting on top of platform \textbf{Drawer:} top of handles or where we stand &
       \centering \textbf{Teller:} small cloud in top center a bit to the right owl is on the platform \textbf{Drawer:} go &
       \centering \textbf{Teller:} there is a medium to small apple tree on right half in blue half in green right side of tree cut off a bit \textbf{Drawer:} go &
       \centering \textbf{Teller:} a dog directly under tree facing left \textbf{Drawer:} size &
       \centering \textbf{Teller:} a girl standing in front of slide arms up smiling dog looks small 
   \end{tabu}\vspace*{5mm}

   \caption{Random selection of examples generated by our Aux model for the CoDraw dataset.}
   \label{fig:random_examples_aux}
\end{figure*}

\begin{figure*}[!h]
   \scriptsize
   \begin{tabu} to \textwidth {XXXXX}
       \centering \includegraphics[scale=0.5]{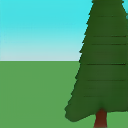} &
       \centering \includegraphics[scale=0.5]{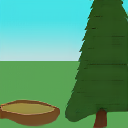} &
       \centering \includegraphics[scale=0.5]{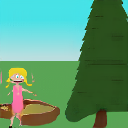} &
       \centering \includegraphics[scale=0.5]{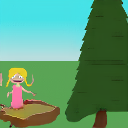} &
       \centering \includegraphics[scale=0.5]{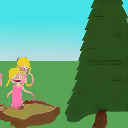} \\
       \centering \textbf{Teller:} large pine tree on right trunk 1 4 inch from bottom cut off top and right &
       \centering \textbf{Teller:} large sand box on left mound on left half bottom edge cut off and left corner cut off mound fully visible &
       \centering \textbf{Teller:} middle of sandbox large girl running right big toothy smile &
       \centering \textbf{Teller:} bottom left corner of sandbox large cat looking right top right corner sandbox is baseball &
       \centering \textbf{Teller:} girl is wearing pirate cap will check when ready 
   \end{tabu}

   \scriptsize
   \begin{tabu} to \textwidth {XXXXX}
       \centering \includegraphics[scale=0.5]{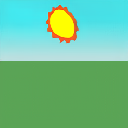} &
       \centering \includegraphics[scale=0.5]{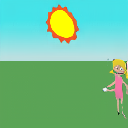} &
       \centering \includegraphics[scale=0.5]{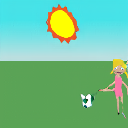} &
       \centering \includegraphics[scale=0.5]{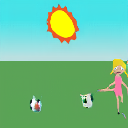} &
       \centering \includegraphics[scale=0.5]{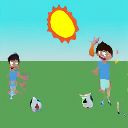} \\
       \centering \textbf{Teller:} there is a large sun in the left hand corner &
       \centering \textbf{Teller:} there is a small girl facing left she has one leg in the air 3 inches from the right 1 2 an inch from the bottom &
       \centering \textbf{Teller:} small duck facing girl bill at her foot like she is going to kick the duck but move the duck 1 4 inch away about an inch from bottom &
       \centering \textbf{Teller:} soccer ball to the left of the duck to the left 1 4 inch and up 2 inches &
       \centering \textbf{Teller:} boy like he is running facing the duck and girl they playing keep away from duck his arms up and looks like he is running mouth open \textbf{Teller:} he has glasses on 
   \end{tabu}

   \scriptsize
   \begin{tabu} to \textwidth {XXXXX}
       \centering \includegraphics[scale=0.5]{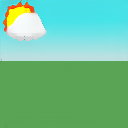} &
       \centering \includegraphics[scale=0.5]{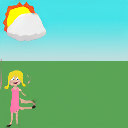} &
       \centering \includegraphics[scale=0.5]{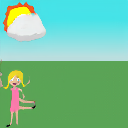} &
       \centering \includegraphics[scale=0.5]{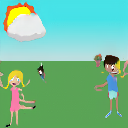} &
       \centering \includegraphics[scale=0.5]{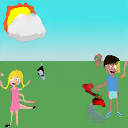} \\
       \centering \textbf{Teller:} top left corner is large sun half of it hidden \textbf{Drawer:} what else \textbf{Teller:} top right corner is large cloud half of it hidden \textbf{Drawer:} &
       \centering \textbf{Teller:} on left side of screen is a large girl standing with her arms in front of her facing right her eyes are even with the horizon line \textbf{Drawer:} &
       \centering \textbf{Teller:} girl is wearing a chefs hat sunglasses and is holding a pink shovel in her right hand \textbf{Drawer:} &
       \centering \textbf{Teller:} on right side of screen is large boy one hand on hip facing the girl with angry expression his neck is even with horizon line \textbf{Drawer:} got it what else &
       \centering \textbf{Teller:} between them is a large grill \textbf{Drawer:} 
   \end{tabu}

   \scriptsize
   \begin{tabu} to \textwidth {XXXXX}
       \centering \includegraphics[scale=0.5]{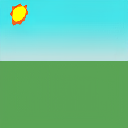} &
       \centering \includegraphics[scale=0.5]{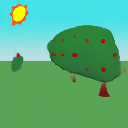} &
       \centering \includegraphics[scale=0.5]{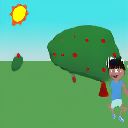} &
       \centering \includegraphics[scale=0.5]{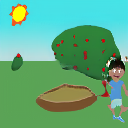} &
       \centering \includegraphics[scale=0.5]{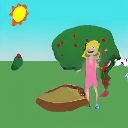} \\
       \centering \textbf{Teller:} big sun on left only top cut a little bit the yellow part &
       \centering \textbf{Teller:} on right sun small apple tree cut top &
       \centering \textbf{Teller:} on right of tree happy mike hand front stand facing left head touches the tree &
       \centering \textbf{Teller:} below sun small sand mike is medium box mound on left left corner is hidden a small duck inside the sandbox &
       \centering \textbf{Teller:} medium jenny sits on right side of sandbox crossed legs and one hand up facing right head cover right corner of sandbox 
   \end{tabu}

   \scriptsize
   \begin{tabu} to \textwidth {XXXXX}
       \centering \includegraphics[scale=0.5]{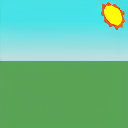} &
       \centering \includegraphics[scale=0.5]{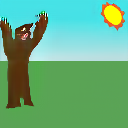} &
       \centering \includegraphics[scale=0.5]{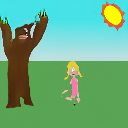} &
       \centering \includegraphics[scale=0.5]{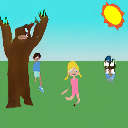} &
       \centering \includegraphics[scale=0.5]{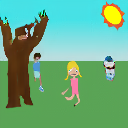} \\
       \centering \textbf{Teller:} sun top right almost touching top 1 `` from left medium size &
       \centering \textbf{Teller:} big bear at left facing right arm pits at horizon &
       \centering \textbf{Teller:} medium to small size girl about 3 `` away from bear facing left angry arms out holding a bat in right hand head in the blue area &
       \centering \textbf{Teller:} boy behind her about 1 5 `` away arms up startled glove on left hand rainbow hat on his eyes slightly below horizon &
       \centering \textbf{Teller:} move boy down about an inch face him to left and his striped hat and we are good 
   \end{tabu}\vspace*{5mm}

   \caption{Random selection of examples generated by our $D$ Concat model for the CoDraw dataset.}
   \label{fig:random_examples_d-concat}
\end{figure*}

\begin{figure*}[!h]
   \scriptsize
   \begin{tabu} to \textwidth {XXXXX}
       \centering \includegraphics[scale=0.5]{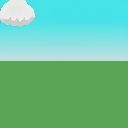} &
       \centering \includegraphics[scale=0.5]{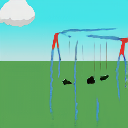} &
       \centering \includegraphics[scale=0.5]{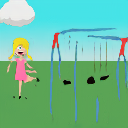} &
       \centering \includegraphics[scale=0.5]{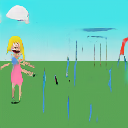} &
       \centering \includegraphics[scale=0.5]{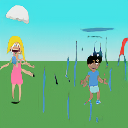} \\
       \centering \textbf{Teller:} big cloud top left side \textbf{Drawer:} got it &
       \centering \textbf{Teller:} on the right side is a swing big size \textbf{Drawer:} any parts cut off &
       \centering \textbf{Teller:} girl on the left side neg horizon one arm up facing right happy face \textbf{Drawer:} got it &
       \centering \textbf{Teller:} one part cut from swing just a bit from the right \textbf{Drawer:} \textbf{Teller:} next to the right of the cloud is a basketball a bit over the cloud \textbf{Drawer:} &
       \centering \textbf{Teller:} a boy is on the swing the right sit legs cross surprised face facing left color hat baseball glove \textbf{Drawer:} got it 
   \end{tabu}

   \scriptsize
   \begin{tabu} to \textwidth {XXXXX}
       \centering \includegraphics[scale=0.5]{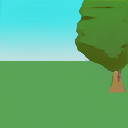} &
       \centering \includegraphics[scale=0.5]{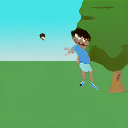} &
       \centering \includegraphics[scale=0.5]{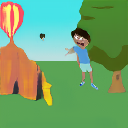} &
       \centering \includegraphics[scale=0.5]{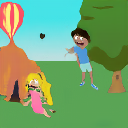} &
       \centering \includegraphics[scale=0.5]{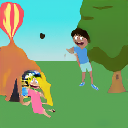} \\
       \centering \textbf{Drawer:} ready \textbf{Teller:} big oak on right hole facing right hole almost touching horizon \textbf{Drawer:} if its large it is huge how much is cut off on the right &
       \centering \textbf{Teller:} smiling big hands out mike on right left trunk point touching his back hair above horizon &
       \centering \textbf{Teller:} small hot balloon on left 1 4 from top 1 in from left big tent on left facing right cut off slightly on back top above horizon &
       \centering \textbf{Teller:} smiling big hands out jenny is in front of left opening tent &
       \centering \textbf{Teller:} she has an owl sitting on her left wrist her hand is in the dark opening \textbf{Drawer:} she is facing him and large owl right 
   \end{tabu}

   \scriptsize
   \begin{tabu} to \textwidth {XXXXX}
       \centering \includegraphics[scale=0.5]{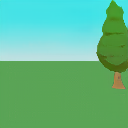} &
       \centering \includegraphics[scale=0.5]{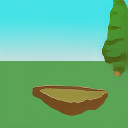} &
       \centering \includegraphics[scale=0.5]{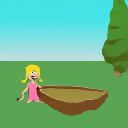} &
       \centering \includegraphics[scale=0.5]{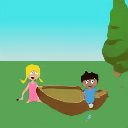} &
       \centering \includegraphics[scale=0.5]{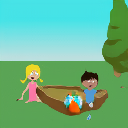} \\
       \centering \textbf{Drawer:} go \textbf{Teller:} small bushy tree facing left owl on right middle of tree &
       \centering \textbf{Teller:} to the left of tree is a medium sandbox mound on right close to bottom \textbf{Drawer:} next &
       \centering \textbf{Teller:} girl sitting in left corner indian style smiling one arm up \textbf{Drawer:} next &
       \centering \textbf{Teller:} boy in right corner sitting indian style smiling with arms open both facing right \textbf{Drawer:} next &
       \centering \textbf{Teller:} under boys right hand is cup in sand straw to left to left of cup is medium beach ball 
   \end{tabu}

   \scriptsize
   \begin{tabu} to \textwidth {XXXXX}
       \centering \includegraphics[scale=0.5]{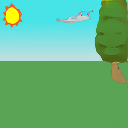} &
       \centering \includegraphics[scale=0.5]{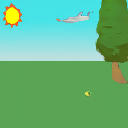} &
       \centering \includegraphics[scale=0.5]{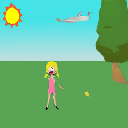} &
       \centering \includegraphics[scale=0.5]{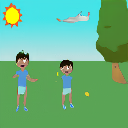} &
       \centering \includegraphics[scale=0.5]{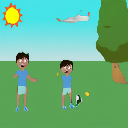} \\
       \centering \textbf{Teller:} tree hole facing left cut off from right side a little bit top hiding a bit of the sun &
       \centering \textbf{Teller:} bumblebee with ear touching the bottom left of tree trunk facing to the right side \textbf{Drawer:} got it &
       \centering \textbf{Teller:} girl sitting smiling facing right hand behind her one inch from side wearing crown &
       \centering \textbf{Teller:} crown almost touches the horizon \textbf{Drawer:} got it \textbf{Teller:} boy faces girl sitting smiling his feet r half inch from hers and raised up a little he wears a beanie with top of it just at horizon &
       \centering \textbf{Teller:} duck between the two with ducks feet level to boys top foot 
   \end{tabu}

   \scriptsize
   \begin{tabu} to \textwidth {XXXXX}
       \centering \includegraphics[scale=0.5]{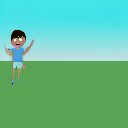} &
       \centering \includegraphics[scale=0.5]{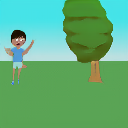} &
       \centering \includegraphics[scale=0.5]{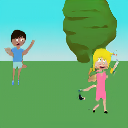} &
       \centering \includegraphics[scale=0.5]{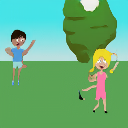} &
       \centering \includegraphics[scale=0.5]{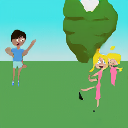} \\
       \centering \textbf{Teller:} on the left an inch from the edge is a boy \textbf{Drawer:} what is he doing \textbf{Teller:} he is facing right standing one hand up teeth showing and holding a racket with one hand that is in the air &
       \centering \textbf{Teller:} next to him a medium tree hole facing left head touches the tip of last branch truck aligns with his waist &
       \centering \textbf{Teller:} on the right 1 inch from edge is a girl sad looking left one hand in the air head aligns with the boy 's \textbf{Drawer:} her left hand cut off &
       \centering \textbf{Teller:} above her is a small cloud right above her can i check &
       \centering \textbf{Teller:} no about 2 cm from the edge the hand \textbf{Teller:} move her 1 more cm from the edge she is holding a yellow small ball in the hand in air 
   \end{tabu}\vspace*{5mm}

   \caption{Random selection of examples generated by our $D$ Subtract model for the CoDraw dataset.}
   \label{fig:random_examples_d-subtract}
\end{figure*}

\begin{figure*}[!h]
  \scriptsize
  \setlength{\tabulinesep}{3pt}
   \begin{tabu} to \textwidth {X[0.3,m]X[1,m]}
       \centering \includegraphics[scale=0.5]{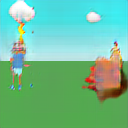} &
     \textbf{Drawer:} what is there \textbf{Teller:} big thunderbolt on left bolt facing right touching on left edge close to top \textbf{Drawer:} and \textbf{Teller:} big raindrop cloud to the right of thunderbolt cloud cutting it off on right side a little big shocked jenny with arms up \textbf{Drawer:} where is she \textbf{Teller:} head right below horizon sad big sitting mike with legs facing right is beside her \textbf{Drawer:} and \textbf{Teller:} he 's wearing a star hat soccer ball is covering his left foot and shin \textbf{Drawer:} and\\
     \centering \includegraphics[scale=0.5]{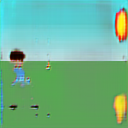} &
     \textbf{Teller:} ready \textbf{Drawer:} and ready \textbf{Teller:} upper right corner large sun with right edges a bit cut off and top cut off \textbf{Teller:} under sun happy boy standing facing left with right arm up his shoulders just above horizon line \textbf{Teller:} he is wearing a pirate hat it touches on of the sun tips on the left side \textbf{Teller:} happy girl kicking on left side she is about 1 5 inches in from left side her mouth is at the horizon line \textbf{Drawer:} got it \textbf{Teller:} just a tiny bit off of girls kicking foot is a beach ball a cloud is over the girl towards the right center \textbf{Drawer:} is the cloud on the right or the sun you said sun upper right corner\\
     \centering \includegraphics[scale=0.5]{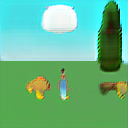} &
     \textbf{Teller:} boy left side kicking leg facing right his half torso aligns with horizon he is shocked \textbf{Drawer:} go \textbf{Teller:} finger away from his leg soccer ball its bottom part touches horizon \textbf{Teller:} right side medium tree 1 4 cut off right side and trunk half way in grass with slight cut off as well right side hole facing left \textbf{Drawer:} go \textbf{Teller:} plain cloud top middle top part cut off big size in front of tree dog its legs behind completely cut off and it 's facing left \textbf{Teller:} near dog is a big cat its tail cover 's dog 's front leg slightly and facing right and then girl sitting smiling facing right\\
     \centering \includegraphics[scale=0.5]{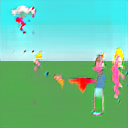} &
     \textbf{Drawer:} ready \textbf{Teller:} top left facing left one 1 4 inch from side blade touch top small helicopter facing left \textbf{Drawer:} what 's in the left the helicopter \textbf{Teller:} nothing it is a 1 4 inch from side flying left \textbf{Drawer:} got it it 's tiny right \textbf{Teller:} yes \textbf{Teller:} below copter is large boy facing right arms out mouth open neck at horizon \textbf{Teller:} right of boy his top hand is on first plank is a large picnic table left top corner is highest point pie is there in corner \textbf{Drawer:} which side is the pie \textbf{Teller:} right of pie is large girl facing left standing with smile no teeth one arm up and one down pie top left corner \textbf{Drawer:} where is she to the horizon and she is in front of the table \textbf{Teller:} girl in front of table nose at horizon top right corner of table is ketchup \textbf{Drawer:} got it \textbf{Teller:} 1 2 inch from right side and 1 2 inch from horizon is large grill\\
     \centering \includegraphics[scale=0.5]{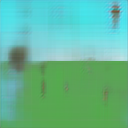} &
       \textbf{Drawer:} ready \textbf{Teller:} there is a medium in the center of the sky just below the top edge \textbf{Drawer:} medium cloud \textbf{Teller:} oh sorry medium sun the medium cloud is down and to the right in the sky \textbf{Teller:} there is a small oak tree on the left an inch away from the left edge hole facing right 2 3s of the leaves are above the horizon \textbf{Teller:} on the right side the kids are both medium sized and facing left jenny is happy jumping half inch from the right edge
   \end{tabu}

   \caption{Random selection of examples generated by our Non-iterative model for the CoDraw dataset.}
   \label{fig:random_examples_non-iterative}
\end{figure*}

We present selected examples generated using our best model ($D$ Subtract) on two datasets. Examples generated for \gls{codraw} are presented in Figure~\ref{fig:codraw_examples} and examples generated for \gls{i-clevr} are presented in Figure~\ref{fig:clevr_examples}. We also present random examples from all the models present in the ablation study for a qualitative comparison on the \gls{codraw} dataset. These are shown in Figure~\ref{fig:random_examples_baseline}~(Baseline), Figure~\ref{fig:random_examples_mismatch}~(Mismatch), Figure~\ref{fig:random_examples_g-prior}~($G$ prior), Figure~\ref{fig:random_examples_aux}~(Aux), Figure~\ref{fig:random_examples_d-concat}~($D$ Concat), Figure~\ref{fig:random_examples_d-subtract}~($D$ Subtract), and Figure~\ref{fig:random_examples_non-iterative}~(Non-iterative).

\section{Generalization to new background images}
\begin{figure*}
  \hfill
  \begin{subfigure}[b]{0.32\textwidth}
     \centering
     \includegraphics[height=0.45\textwidth,width=0.45\textwidth]{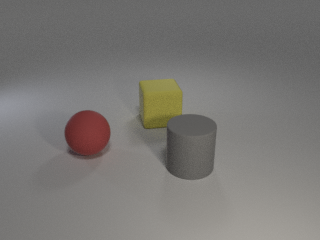}
     \includegraphics[height=0.45\textwidth]{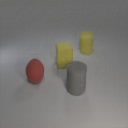}
     \caption[]%
     {{\small \textbf{Left:} Initial Image \textbf{Right:} Final Image \textbf{Instruction:} Add a yellow cylinder behind the gray cylinder on the right and behind the yellow cube on the right}}
  \end{subfigure}
  \hfill
  \begin{subfigure}[b]{0.32\textwidth}
     \centering
     \includegraphics[height=0.45\textwidth,width=0.45\textwidth]{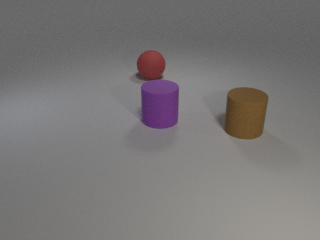}
     \includegraphics[height=0.45\textwidth]{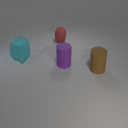}
     \caption[]%
     {{\small \textbf{Left:} Initial Image \textbf{Right:} Final Image \textbf{Instruction:} Add a cyan cube behind the brown cylinder on the left and behind the purple cylinder on the left}}
  \end{subfigure}
  \hfill
  \begin{subfigure}[b]{0.32\textwidth}
     \centering
     \includegraphics[height=0.45\textwidth,width=0.45\textwidth]{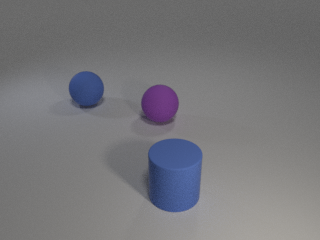}
     \includegraphics[height=0.45\textwidth]{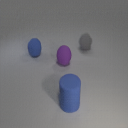}
     \caption[]%
     {{\small \textbf{Left:} Initial Image \textbf{Right:} Final Image \textbf{Instruction:} Add a gray sphere behind the blue cylinder on the right and behind the purple sphere on the right}}
  \end{subfigure}\vspace*{5mm}
  \hfill
  \caption[]%
  {\small When GeNeVA-GAN is provided with an initial image different from the background image used during training, it still adds the desired object with the right properties at the correct location. The model was not trained in this setting and the success of this experiment demonstrates that it has learnt to preserve the existing canvas, understand the existing objects, and add new objects with the correct relationships to existing objects.}
  \label{fig:intermediate}
\end{figure*}

GeNeVA-GAN was trained using the empty background image as the initial image. We ran an experiment where we used a different image (intermediate ground truth image from the test set containing objects) as the initial image. We present generated examples from this experiment in Figure~\ref{fig:intermediate}. The model is able to place the desired object at the correct location with the correct color and shape over the provided image. This shows that the model is capable of generalizing to a background it was not trained on and it can understand the existing objects from just the initial image without any instruction history for placing them.

\section{\gls{i-clevr} Dataset Generation}

To generate the image for each step in the sequence, an object with random
attributes is rendered to the scene using Blender \cite{blender}. We
ensure that all objects have a unique combination of attributes. Each object
can have one of 3 shapes (cube, sphere, cylinder) and  one of 8 colors. In
contrast to CLEVR, we have a fixed material and size for objects. For the first
image in the sequence, the object placement is fixed to the image center. For
all the following images, the objects are placed in a random position while
maintaining visibility (not completely occluded) and at a minimum distance from
other objects.

To generate instructions, we use a simple text template that depends on the instruction number. For example,
the second instruction in the sequence will have the following template:

\vspace*{2mm}
\textit{``Add a [object color] [object shape] [relative position: depth] it on the [relative position: horizontal]''}
\vspace*{2mm}

From the third instruction onward, the object position is described relative
to two objects. These two objects are chosen randomly from the existing
objects in the scene.

\section{Additional implementation details}
We use 300-dimensional GloVe\footnote{\scriptsize \url{http://nlp.stanford.edu/data/glove.840B.300d.zip}}
word embeddings for representing the words in each instruction $q_t$. These word embeddings are encoded
using a bi-directional-GRU to obtain a 1024-dimensional instruction encoding $d_t$. All state dimensions
for the higher level GRU $R$ are set to 1024. The output of the conditioning augmentation module is also
1024-dimensional.

The code for this project was implemented in PyTorch~\cite{pytorch}.
For the generator and discriminator optimizers, ``betas'' was set to $(0.0, 0.9)$ and weight decay was set to $0$.
The learning rates for the image encoding modules were set to $0.006$.
Gradient norm was clipped at 50. For each training experiment,
we used a batch size of $32$ over $2$ NVIDIA P100 GPUs.

\section{Additional language encoder experiments}
We experimented with using skip-thought encoding for sentences
instead of training the bi-directional-GRU encoder over GloVe embeddings.
For the paper, we chose to use the latter since it performed better.

We also experimented with passing the previous image through
the language encoder, but observed that it was easier for the model
to generate an accurate image when the previous image features are passed
to the Generator directly.

